\documentclass{llncs}

\usepackage{graphicx} 
\usepackage{subcaption} 

\usepackage{algorithm}
\usepackage{algorithmic}
\usepackage{amsmath}

\usepackage{tikz}

\usepackage{hyperref}

\begin{document}
\title{Building Ensembles of Adaptive Nested Dichotomies with Random-Pair Selection}
	
\author{Tim Leathart \and Bernhard Pfahringer \and Eibe Frank}
\institute{Department of Computer Science, University of Waikato, New Zealand \\ 
		   \texttt{tml15@students.waikato.ac.nz, \{bernhard,eibe\}@cs.waikato.ac.nz}}
	
\maketitle

\begin{abstract} 
A system of nested dichotomies is a method of decomposing a multi-class problem into a collection of binary problems. Such a system recursively applies binary splits to divide the set of classes into two subsets, and trains a binary classifier for each split. Although ensembles of nested dichotomies with random structure have been shown to perform well in practice, using a more sophisticated class subset selection method can be used to improve classification accuracy. We investigate an approach to this problem called random-pair selection, and evaluate its effectiveness compared to other published methods of subset selection. We show that our method outperforms other methods in many cases when forming ensembles of nested dichotomies, and is at least on par in all other cases.
\end{abstract} 

\section{Introduction}
Multi-class classification problems -- problems with more than two classes -- are commonplace in real world scenarios. Some learning methods can handle multi-class problems inherently, \textit{e.g.}, decision tree inducers, but others may require a different approach. Even techniques such as decision tree inducers may benefit from methods that decompose a multi-class problem in some manner. Typically, a collection of binary classifiers is trained and combined in some way to produce a multi-class classification. This process is called binarization. Popular techniques for adapting binary classifiers to multi-class problems include pairwise classification~\cite{hastie1998classification}, one-vs-all classification~\cite{rifkin2004defense}, and error correcting output codes~\cite{dietterich1995solving}. Ensembles of nested dichotomies~\cite{frank2004ensembles} have been shown to be an effective substitute to these methods. Depending on the base classifier used, they can outperform both pairwise classification and error-correcting output codes~\cite{frank2004ensembles}. 

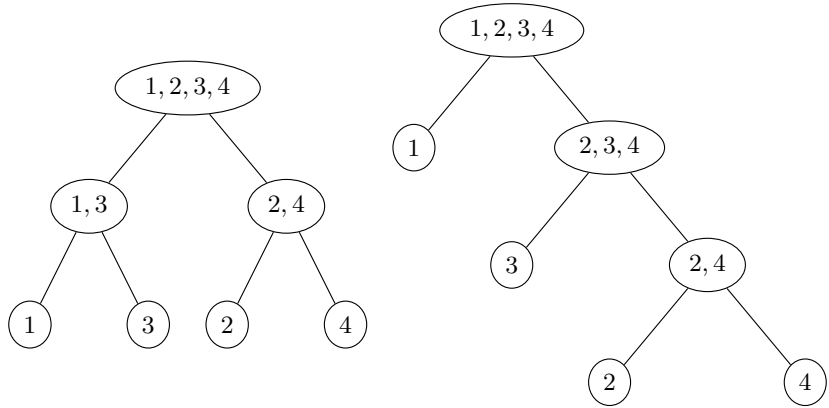
\begin{figure*}[t]
	\centering
	\begin{subfigure}{5cm}
	\resizebox{5cm}{!}{
	\begin{tikzpicture}
		\tikzstyle{level 1}=[sibling distance = 25mm]
		\tikzstyle{level 2}=[sibling distance = 15mm]
		\usetikzlibrary{shapes}
		\node[ellipse,draw](z){${1,2,3,4}$}
		  child
		  {
		  	node[ellipse,draw]{$1,3$}
		  	child
		  	{
		  		node[ellipse,draw]{$1$}
		  	}
		  	child
		  	{
		  		node[ellipse,draw]{$3$}
		  	}
		  }
		  child
		  {
	    		node[ellipse,draw]{$2,4$} 
	    		child
	    		{
	    			node[ellipse,draw] {$2$}
	    		} 
	    		child
	    		{
	    			node[ellipse,draw] {$4$}
	    		}
	    	  };
	\end{tikzpicture}
	}
	\end{subfigure}
	\begin{subfigure}{6cm}
	\resizebox{6cm}{!}{
	\begin{tikzpicture}
		\tikzstyle{level 1}=[sibling distance = 25mm]
		\usetikzlibrary{shapes}
		\node[ellipse,draw](z){${1,2,3,4}$}
		  child
		  {
		  	node[ellipse,draw]{$1$}
		  }
		  child
		  {
	    		node[ellipse,draw]{$2,3,4$} 
	    		child
	    		{
	    			node[ellipse,draw] {$3$}
	    		} 
	    		child
	    		{
	    			node[ellipse,draw] {$2,4$}
	    			child
	    			{
	    				node[ellipse,draw] {$2$}
	    			}
	    			child
	    			{
	    				node[ellipse,draw] {$4$}
	    			}
	    		}
	    	  };
	\end{tikzpicture}
	}
	\end{subfigure}
	\caption{\label{fig:dichotomy_example} Two examples of nested dichotomies for a four class problem.}
\end{figure*}

In a nested dichotomy, the set of classes is split into two subsets recursively until there is only one class in each subset. Nested dichotomies are represented as binary tree structures (Fig.~\ref{fig:dichotomy_example}). At each node of a nested dichotomy, a binary classifier is learned to classify instances as belonging to one of the two subsets of classes. A nice feature of nested dichotomies is that class probability estimates can be computed in a natural way if the binary classifier used at each node can output two-class probability estimates. 

The number of nested dichotomies for a $c$-class problem increases exponentially with the number of classes. One approach is to sample nested dichotomies at random to form an ensemble of them~\cite{frank2004ensembles}. However, this may result in binary problems that are difficult to learn for the base classifier.

This paper is founded on the observation that some classes are generally easier to separate than others. For example, in a dataset of images of handwritten digits, the digits `5' and `6' are are much more difficult to distinguish than the digits `0' and `1'. This means that if `5' and `6' were put into opposite class subsets, the base classifier would have a more difficult task to discriminate the two subsets than if they were grouped together. Moreover, if the base classifier assigns high probability to an incorrect branch when classifying a test instance, it is unlikely that the final prediction will be correct. Therefore, we should try to group similar classes into the same class subsets whenever possible, and separate them in lower levels of the tree near the leaf nodes. 

In this paper, we propose a method for semi-random class subset selection, which we call ``random-pair selection'', that attempts to group similar classes together for as long as possible. This means that the binary classifiers close to the root of the tree of classes can learn to distinguish higher-level features, while the ones close to the leaf nodes can focus on the more fine-grained details between similar classes. We evaluate this method against other published class subset selection strategies.

This paper is structured as follows. In Section~\ref{sec:related_work}, we give a review of other adaptations of ensembles of nested dichotomies. In Section~\ref{sec:random_pair}, we describe the random-pair selection strategy and give an overview of how it works. We also cover theoretical advantages of our method over other methods, and give an analysis of how this strategy affects the space of possible nested dichotomy trees to sample from. In Section~\ref{sec:experimental_results}, we evaluate these methods and compare them to other class subset selection techniques. 

\section{\label{sec:related_work}Related Work} 
The original framework of ensembles of nested dichotomies by Frank and Kramer was proposed in 2004~\cite{frank2004ensembles}. In this framework, a binary tree is sampled randomly from the set of possible trees, based on the assumption that each nested dichotomy is equally likely to be useful \textit{a priori}. By building an ensemble of nested dichotomies in this manner, Frank and Kramer achieved results that are competitive with other binarization techniques using decision trees and logistic regression as the two-class models for each node. 

There have been a number of adaptations of ensembles of nested dichotomies since, mainly focusing on different class selection techniques. Dong~\textit{et al.} propose to restrict the space of nested dichotomies to only consist of structures with balanced splits~\cite{dong2005ensembles}. Doing this regulates the depth of the trees, which can reduce the size of the training data for each binary classifier and thus has a positive effect on the runtime. It was shown empirically that this method has little effect on accuracy. Dong~\textit{et al.} also consider nested dichotomies where the number of instances per subset is approximately balanced at each split, instead of the number of classes. This also reduces the runtime, but can aversely effect the accuracy in rare cases. 

The original framework of ensembles of nested dichotomies uses randomization to build an ensemble, \textit{i.e.}, the structure of each nested dichotomy in the ensemble is randomly selected, but built from the same data. Rodriguez~\textit{et al.} explore the use of other ensemble techniques in conjunction with nested dichotomies~\cite{rodriguez2010forests}. The authors found that improvements in accuracy can be achieved by using bagging~\cite{breiman1996bagging}, AdaBoost~\cite{freund1996game} and MultiBoost~\cite{webb2000multiboosting} with random nested dichotomies as the base learner, compared to solely randomizing the structure of the nested dichotomies. The authors also experimented with different base classifiers for the nested dichotomies, and found that using ensembles of decision trees as base classifiers yielded favourable results compared to individual decision trees.

Duarte-Villase{\~n}or~\textit{et al.} propose to split the classes more intelligently than randomly by using various clustering techniques~\cite{duarte2012nested}. They first compute the centroid of each class. Then, at each node of a nested dichotomy, they select the two classes with the furthest centroids as initial classes for each subset. Once the two classes have been picked, the remaining classes are assigned to one of the two subsets based on the distance of their centroids to the centroids of the initial classes. Duarte-Villase{\~n}or~\textit{et al.} evaluate three different distance measures for determining the furthest centroids, taking into account the position of the centroids, the radius of the clusters and average distance of each instance from the centroid. They found that these class subset selection methods gave superior accuracy to the random methods previously proposed when the nested dichotomies were used for boosting.

\section{\label{sec:random_pair}Random-Pair Selection}
We present a class selection strategy for choosing subsets in a nested dichotomy called random-pair selection. This has the same intention as the centroid-based methods proposed by Duarte-Villase{\~n}or~\textit{et al.}~\cite{duarte2012nested}. Our method differs in that it takes a more direct approach to discovering similar classes by using the actual base classifier to decide which classes are more easily separable. Moreover, it incorporates an aspect of randomization.

\subsection{The Algorithm}
The process for constructing a nested dichotomy with random-pair selection is as follows:
\begin{enumerate}
\item Create a root node for the tree.
\item If the class set $C$ has only one class, then create a leaf node.
\item Otherwise, split $C$ into two subsets by the following:
\begin{enumerate}
\item Select a pair of classes $c_1, c_2 \in C$ at random, where $C$ is the set of all classes present at the current node. 
\item Train a binary classifier using these two classes as training data. Then, use the remaining classes as test data, and observe which of the initial classes the majority of instances of each test class are classified as.\footnote{When the dataset is large, it may be sensible to subsample the training data at each node when performing this step.}
\item Two subsets are created, using the initial classes: $s_1 = \left\{c_1\right\}, s_2 = \left\{c_2\right\}$
\item The test classes $c_n \in C \setminus \left\{c_1, c_2\right\}$ are added to $s_1$ or $s_2$ based on whether $c_n$ is more likely to be classified as $c_1$ or $c_2$. 
\item A new binary model is trained using the full data at the node, using the new class labels $s_1$ and $s_2$ for each instance.
\end{enumerate}
\item Create new nodes for both $s_1$ and $s_2$ and recurse for each child node from Step~2.
\end{enumerate}

\begin{figure*}[t]
	\centering
	\begin{subfigure}{2.36cm}
		\includegraphics[width=2.36cm]{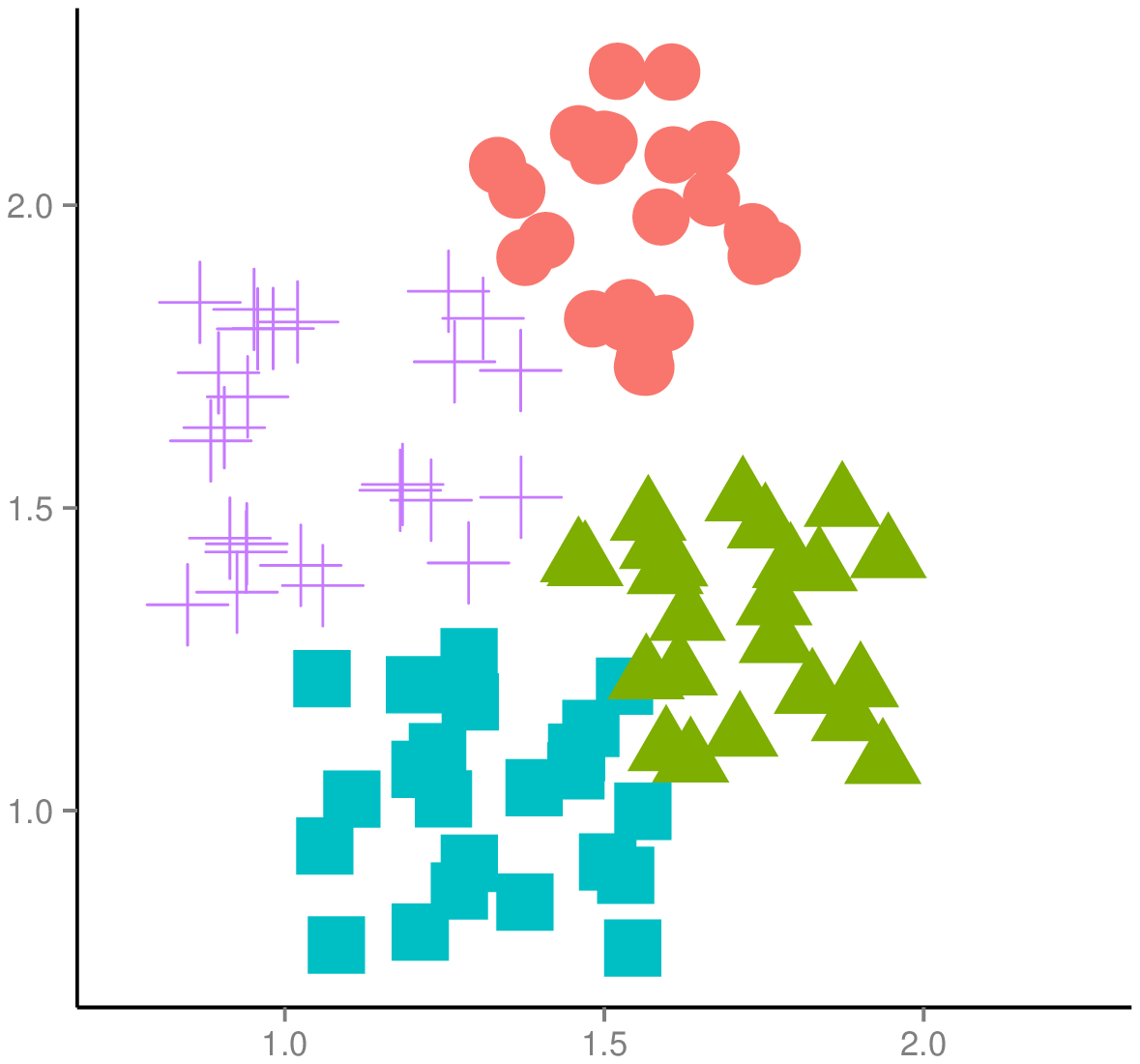}
		\caption{}
	\end{subfigure}
	\begin{subfigure}{2.36cm}
		\includegraphics[width=2.36cm]{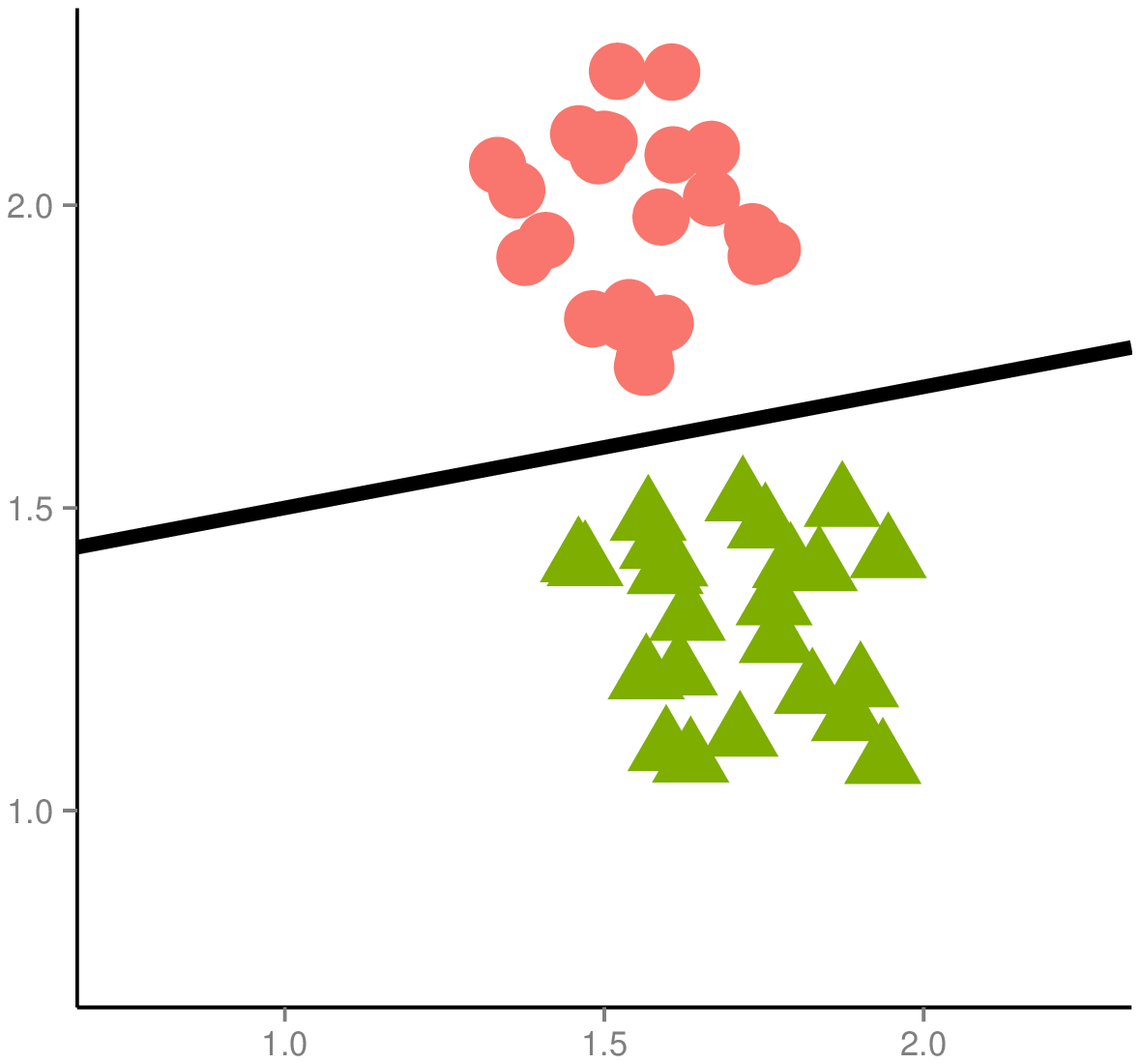}
		\caption{}
	\end{subfigure}
	\begin{subfigure}{2.36cm}
		\includegraphics[width=2.36cm]{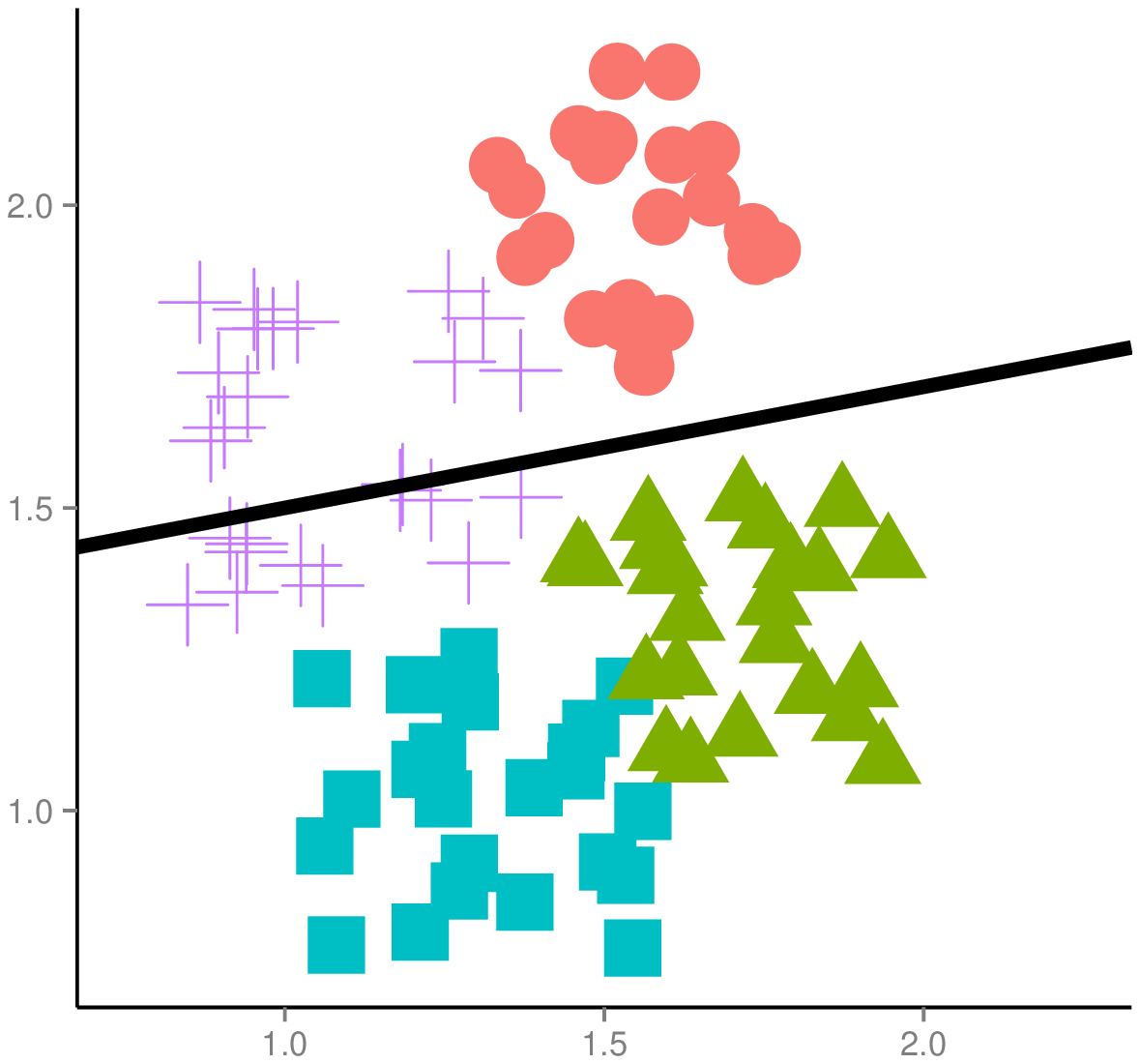}
		\caption{}
	\end{subfigure}
	\begin{subfigure}{2.36cm}
		\includegraphics[width=2.36cm]{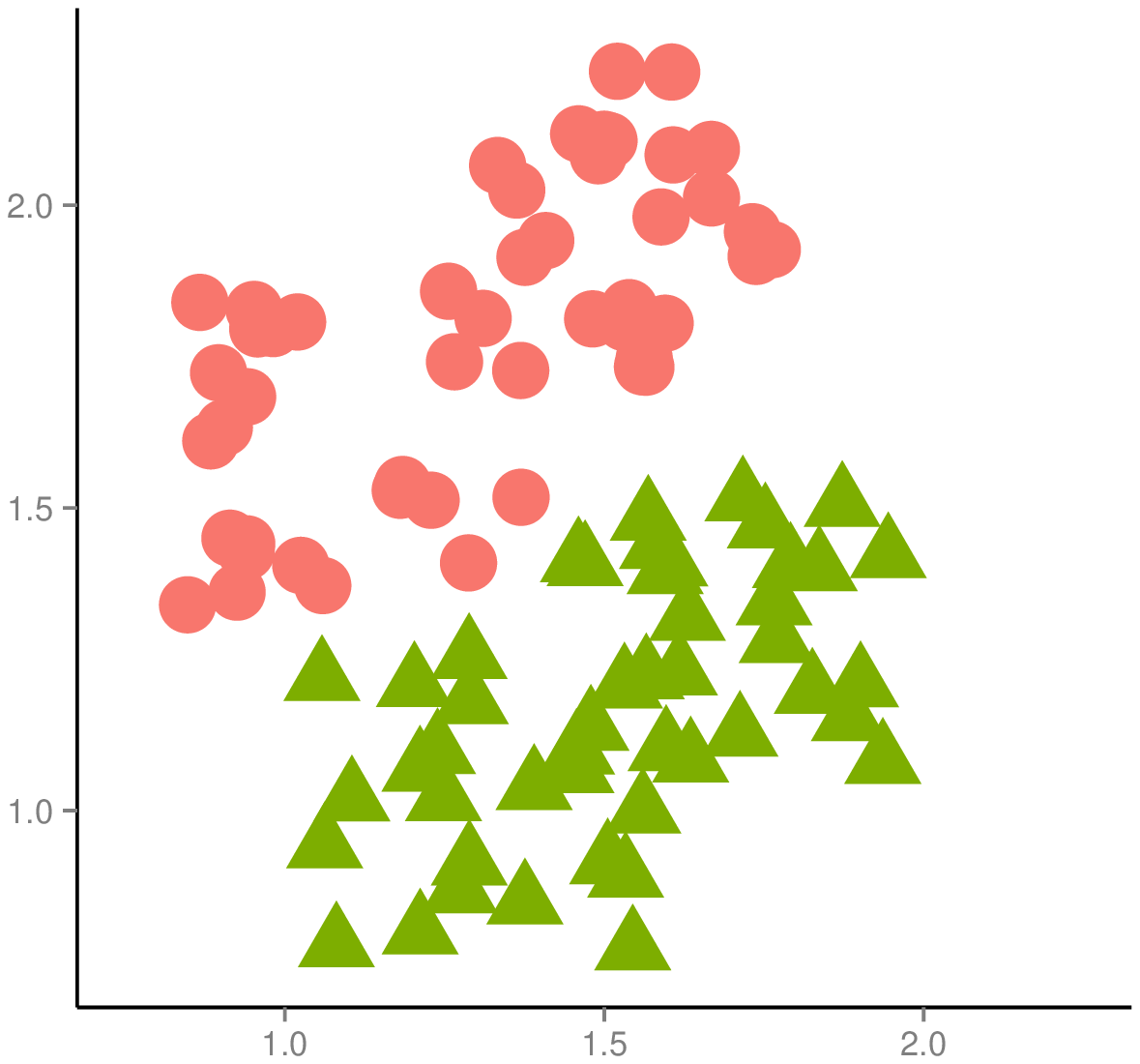}
		\caption{}
	\end{subfigure}
	\begin{subfigure}{2.36cm}
		\includegraphics[width=2.36cm]{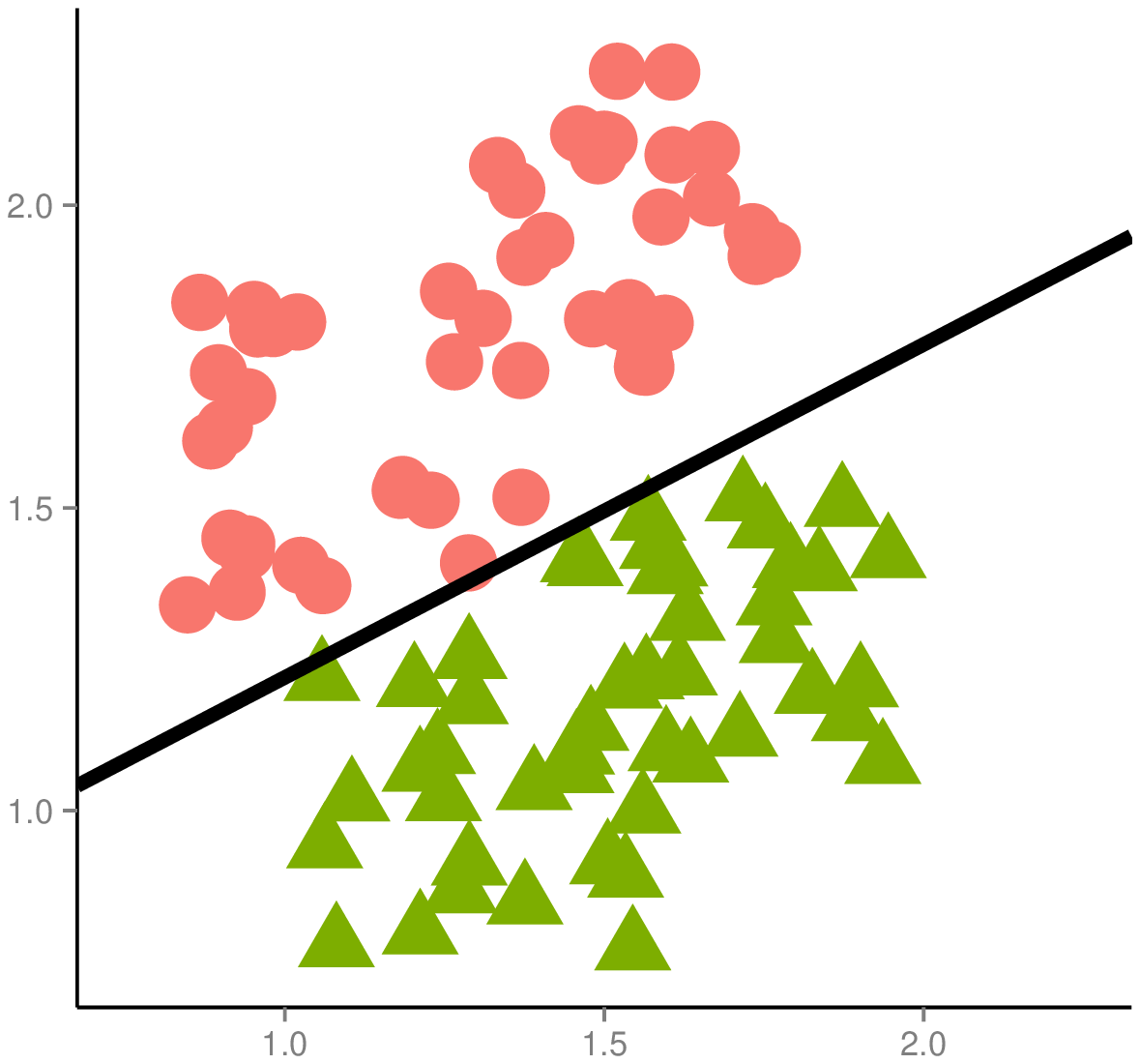}
		\caption{}
	\end{subfigure}
	\caption{\label{fig:randompair}Random-Pair Selection. (a) Original multi-class data. (b) Two classes are selected at random, and a binary classifier is trained on this data. (c) The binary classifier is tested on the other classes. The majority of the `plus' class is classified as `circle', and all of the `square' class is classified as `triangle'. (d)~Combine the classes into subsets based on which of the original classes each new class is more likely to be classified as. (e) Learn another binary classifier, which will be used in the final nested dichotomy tree.}
\end{figure*}

This selection algorithm is illustrated in Fig.~\ref{fig:randompair}. The process for making predictions when using this class selection method is identical to the process for the original ensembles of nested dichotomies. Assuming that the base classifier can produce class probability estimates, the probability of an instance belonging to a class is the product of the estimates given by the binary classifiers on the path from the root to the leaf node corresponding to the particular class.

\subsection{\label{sec:analysis} Analysis of the Space of Nested Dichotomies}
To build an ensemble of nested dichotomies, a set of nested dichotomies needs to be sampled from the space of all nested dichotomies. The size of this space grows very quickly as the number of classes increases. Frank and Kramer calculate that the number of potential nested dichotomies is $(2c-3)!!$ for a $c$-class problem~\cite{frank2004ensembles}. For a 10-class problem, this equates to $34,459,425$ distinct systems of nested dichotomies. Using a class-balanced class-subset selection strategy reduces this number:

\begin{equation}
	T(c) = 
	\begin{cases}
		\frac{1}{2} \binom{c}{c/2} T(\frac{c}{2}) T(\frac{c}{2}), & \text{if } c \text{ is even} \\
		\binom {c}{(c+1)/2} T(\frac{c+1}{2}) T(\frac{c-1}{2}), & \text{if } c \text{ is odd} \\
	\end{cases}
\end{equation}

where $T(2) = T(1) = 1$~\cite{dong2005ensembles}. The number of class-balanced nested dichotomies is still very large, giving $113,400$ possible nested dichotomies for a 10-class problem. The subset selection method based on clustering~\cite{duarte2012nested} takes this idea to the extreme, and gives only a single nested dichotomy for any given number of classes because the class subset selection is deterministic. Even though the system produced by this subset selection strategy is likely to be a useful one, it does not lend itself well to ensemble methods.

The size of the space of nested dichotomies that we sample using the random-pair selection method varies for each dataset, and is dependent on the base classifier. The upper bound for the number of possible binary problems at each node is the number of ways to select two classes at random from a $c$-class dataset,~\textit{i.e.}, $\binom{c}{2}$. In practice, many of these randomly chosen pairs are likely to produce the same class subsets under our method, so the number of possible class splits is likely to be lower than this value. For illustrative purposes, we empirically estimate this value for the logistic regression base learner. We enumerate and count the number of possible class splits for our splitting method at each node of a nested dichotomy for a number of datasets, and plot this number against the number of classes at the corresponding node (Fig.~3a). We also show a similar plot for the case where C4.5 is used as the base classifier (Fig.~3b). Fitting a second degree polynomial to the data for logistic regression yields

\begin{equation}
	p(c) = 0.3812c^2 - 1.4979c + 2.9027.
\end{equation}	

Assuming we apply logistic regression, we can estimate the number of possible class splits for an arbitrary number of classes based on this expression by making a rough estimate of the distribution of classes at each node. Nested dichotomies constructed with random-pair selection are not guaranteed to be balanced, so we average the class subset proportions over a large sample of nested dichotomies on different datasets to find that the two class subsets contain $\frac{1}{3}$ and $\frac{2}{3}$ respectively of the classes on average. Given this information, we can estimate the number of possible nested dichotomies with logistic regression by the recurrence relation

\begin{equation}\label{eqn:rpnd}
	T(c) = p(c) T(\frac{c}{3}) T(\frac{2c}{3})
\end{equation}

where $T(c) = 1$ when $c \leq 2$. Table~\ref{tab:number_of_dichotomies} shows the number of distinct nested dichotomies that can be created for up to 12 classes for the random-pair selection method, class-balanced and completely random selection when we apply this estimate.

\begin{figure*}[t]
	\centering
	\begin{subfigure}{6cm}
		\includegraphics[scale=0.35]{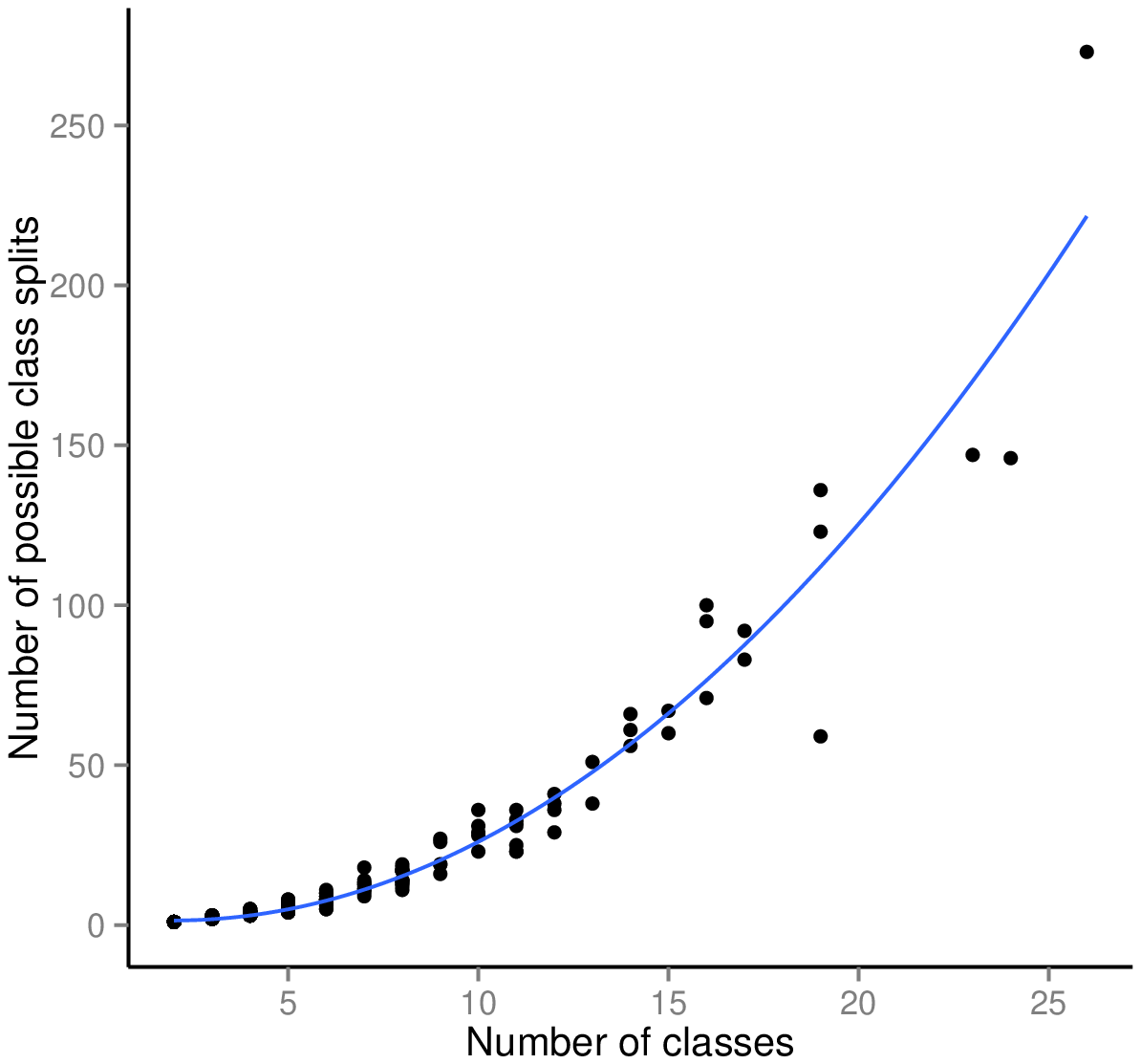}
		\caption{\label{fig:logistic_classes}Logistic regression}
	\end{subfigure}
	\begin{subfigure}{6cm}
		\includegraphics[scale=0.35]{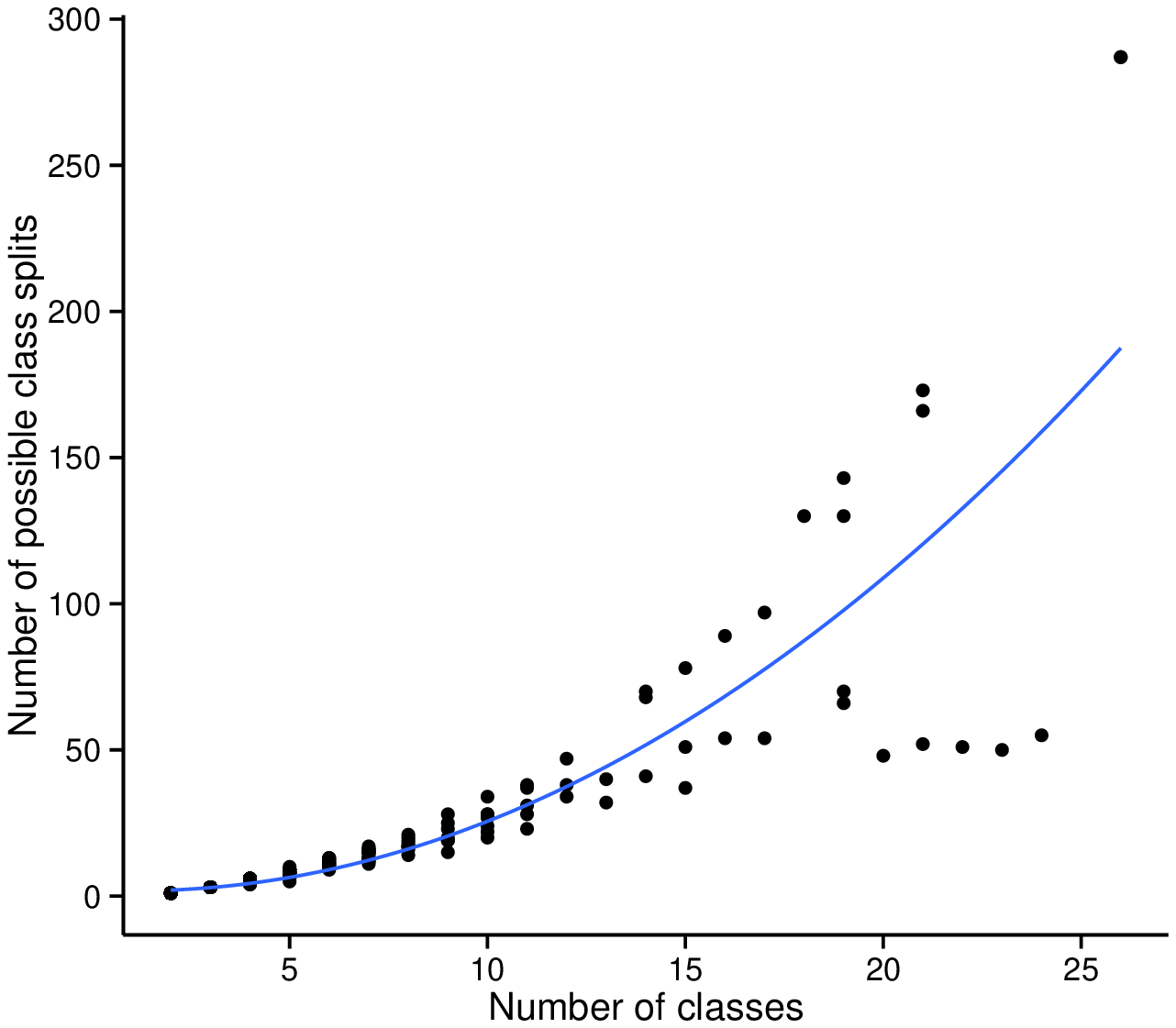}
		\caption{\label{fig:j48_classes}C4.5}
	\end{subfigure}	
	
	\caption{\label{fig:classes_vs_splits} Number of possible splits under a random-pair selection method vs number of classes for a number of UCI datasets.}
\end{figure*}

\begin{table}[t]
\centering
\caption{\label{tab:number_of_dichotomies}The number of possible nested dichotomies for up to 12 classes for each class subset selection technique. The first two columns are taken from~\cite{dong2005ensembles}, and the random-pair column is estimated from (\ref{eqn:rpnd}).}
\scriptsize{
\begin{tabular}{cccc}
\hline
\begin{tabular}[c]{@{}c@{}}Number of \\ classes\end{tabular} & \begin{tabular}[c]{@{}c@{}}Number of \\ nested dichotomies\end{tabular} & \begin{tabular}[c]{@{}c@{}}Number of class-balanced \\ nested dichotomies\end{tabular} & \begin{tabular}[c]{@{}c@{}}Number of random-pair\\ nested dichotomies\end{tabular} \\
\hline
2                      & 1                                                                       & 1                                                                                      & 1\\
3                      & 3                                                                       & 3                                                                                      & 1\\
4                      & 15                                                                      & 3                                                                                      & 5\\
5                      & 105                                                                     & 30                                                                                     & 15\\
6                      & 945                                                                     & 90                                                                                     & 36\\
7                      & 10,395                                                                  & 315                                                                                    & 182\\
8                      & 135,135                                                                 & 315                                                                                    & 470\\
9                      & 2,027,025                                                               & 11,340                                                                                 & 1,254\\
10                     & 34,459,425                                                              & 113,400                                                                                & 7,002\\
11                     & 654,729,075                                                             & 1,247,400                                                                              & 28,189\\
12                     & 13,749,310,575                                                          & 3,742,200                                                                              & 81,451\\
\hline
\end{tabular}
}
\end{table}

\subsection{\label{sec:advantages} Advantages Over Centroid Methods}
Random-pair selection has two theoretical advantages compared to the centroid-based methods proposed by the authors of~\cite{duarte2012nested}: (a) an element of randomness makes it more suitable for ensemble learning, and (b) it adapts to the base classifier that is used.

In the centroid-based methods, each class split is deterministically chosen based on some distance metric. This means that the structure of every nested dichotomy in an ensemble will be the same. This is less important in ensemble techniques that alter the dataset or weights inside the dataset (\textit{e.g.}, bagging or boosting). However, an additional element of randomization in ensembles is typically beneficial. When random-pair selection is employed, the two initial classes are randomly selected in all nested dichotomies, increasing the total number of nested dichotomies that can be constructed as discussed in the previous section.

\begin{figure*}[t]
	\centering
	\begin{subfigure}{10cm}
		\includegraphics[width=\textwidth]{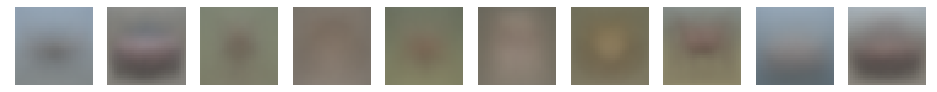}	
	\end{subfigure}
	\begin{subfigure}{10cm}
		\includegraphics[width=\textwidth]{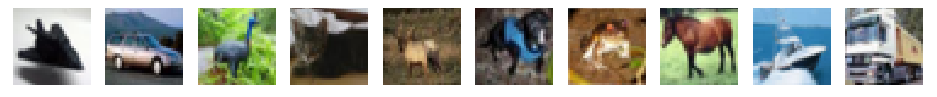}
	\end{subfigure}
	\caption{\label{fig:cifar_centroids}Class centroids of the training component of the CIFAR-10 dataset (above). Samples from each class (below).}
\end{figure*}

Centroid-based methods assume that a smaller distance between two class centroids is indicative of class similarity. While it is true that this is often the case, sometimes the centroids can be relatively meaningless. An example is the CIFAR-10 dataset, a collection of small natural images of various categories such as cats, dogs and trucks~\cite{krizhevsky2009learning}. The classes are naturally divided into two subsets~--~animals and vehicles. Fig.~\ref{fig:cifar_centroids} shows an image representation of the centroids of each class, and a sample image from the respective class below it. It is clear to see that most of these class centroids do not contain much useful information for discriminating between the classes.

\begin{figure*}[t]
	\centering
	\includegraphics[width=8cm]{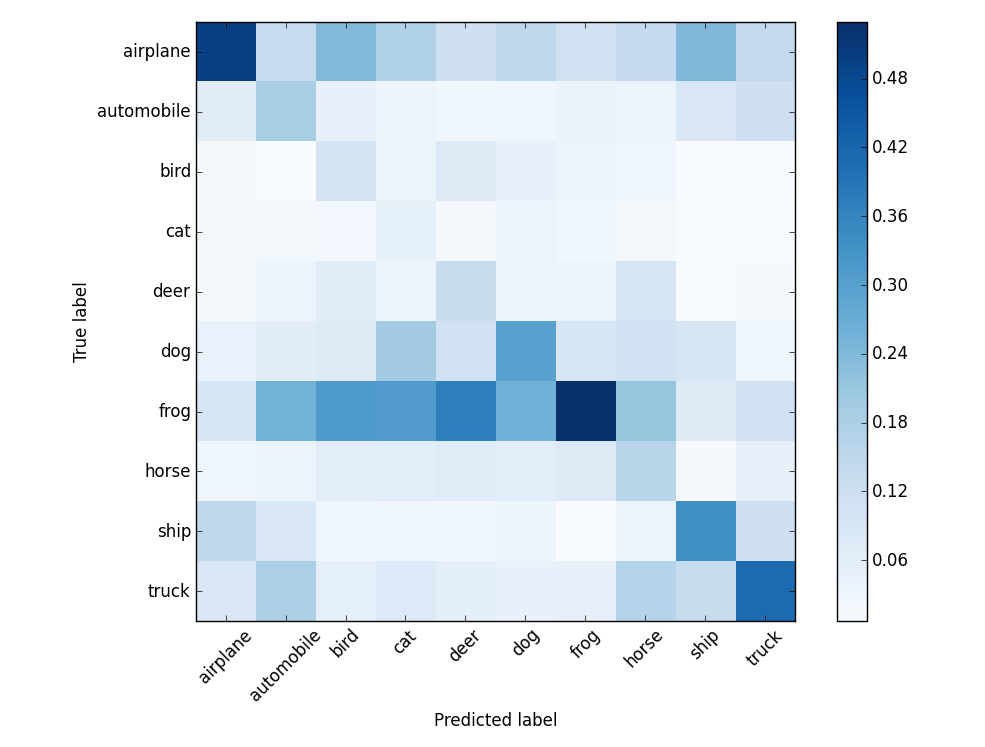}
	\caption{\label{fig:cifar_centroid_confusion}Confusion matrix of a centroid classifier for the CIFAR-10 dataset. The darkness of each square corresponds with the number of instances classified as a particular class.}
\end{figure*}

This effect is clearer when evaluating a simple classifier that classifies instances according to the closest centroid of the training data. For illustrative purposes, see the confusion matrix of such a classifier when trained on the CIFAR-10 dataset (Fig.~\ref{fig:cifar_centroid_confusion}). It is clear to see from the confusion matrix that the centroids cannot be relied upon to produce meaningful predictions in all cases for this data.

A disadvantage of random-pair selection compared to centroid-based methods is an increase in runtime. Under our method, we need to train additional base classifiers during the class subset selection process. However, the extra base classifiers are only trained on a subset of the data at a node,~\textit{i.e.}, only two of the classes, and we can subsample this data during this step if we need to improve the runtime further.

\section{\label{sec:experimental_results}Experimental Results}
We present an evaluation of the random-pair selection method on 18 datasets from the UCI repository~\cite{asuncion2007uci}. Table~\ref{tab:datasets} lists and describes the datasets we used. We specifically selected datasets with at least five classes, as our method will not have a large impact on datasets with few classes. This is due to the fact that there is a relatively small number of possible nested dichotomies for small numbers of classes.

\subsection{Experimental Setup}

All experiments were conducted in WEKA~\cite{hall2009weka}, and performed with 10 times 10-fold cross validation.\footnote{Our implementations can be found in the \texttt{ensemblesOfNestedDichotomies} package in WEKA.} The default settings in WEKA for the base learners and ensemble methods were used in our evaluation. We compared our class subset selection method (RPND) to nested dichotomies based on clustering (NDBC)~\cite{duarte2012nested}, class-balanced nested dichotomies (CBND)~\cite{dong2005ensembles}, and completely random selection (ND)~\cite{frank2004ensembles}. We did not compare against other variants of nested dichotomies such as data-balanced nested dichotomies~\cite{dong2005ensembles}, nested dichotomies based on clustering with radius~\cite{duarte2012nested} and nested dichotomies based on clustering with average radius~\cite{duarte2012nested}, because they were found to either have the same or worse performance on average in~\cite{dong2005ensembles} and~\cite{duarte2012nested} respectively. We used logistic regression and C4.5 as the base learners for our experiments, as they occupy both ends of the bias-variance spectrum. In our results tables, a bullet ($\bullet$) indicates a statistically significant accuracy gain, and an open circle ($\circ$) indicates a statistically significant accuracy reduction ($p=0.05$) by using the random-pair method compared with another method. To establish significance, we used the corrected resampled paired t-test~\cite{nadeau2003inference}.

\begin{table}[t]
\centering
\scriptsize{
\caption{\label{tab:datasets}The datasets used in this evaluation}
	\begin{tabular}{lccc|lccc} \\ 
		\hline
		Dataset & Classes & Instances & Attributes 	&	Dataset & Classes & Instances & Attributes \\
		\hline
		audiology & 24 & 226 & 70 & optdigits & 10 & 5620 & 65\\
		krkopt & 18 & 28056 & 7 &		page-blocks & 5 & 5473 & 11  \\
		LED24 & 10 & 5000 & 25 &		pendigits & 10 & 10992 & 17\\
		letter & 26 & 20000 & 17 & 		segment & 7 & 2310 & 20\\
		mfeat-factors & 10 & 2000 & 217 & 		shuttle & 7 & 58000 & 10\\
		mfeat-fourier & 10 & 2000 & 77 & 		usps & 10 & 9298 & 257\\
		mfeat-karhunen & 10 & 2000 & 65 & 		vowel & 11 & 990 & 14\\
		mfeat-morphological & 10 & 2000 & 7 	&	yeast & 10 & 1484 & 9\\
		mfeat-pixel & 10 & 2000 & 241 & 		zoo & 7 & 101 & 18\\
		\hline	
	\end{tabular}
}
\end{table}


\subsection{Single Nested Dichotomy}
We expect that intelligent class subset selection will have a larger impact in small ensembles of nested dichotomies. This is due to the fact as ensembles grow larger, the worse performing ensemble members will not have as great an influence over the final predictions. Therefore, we first compare a single nested dichotomy using random-pair selection to a single nested dichotomy obtained with other class selection methods. 

\begin{table}[t]
\scriptsize
{\centering 
\caption{\label{tab:single_classifier}Accuracy of a single nested dichotomy with (a) logistic regression and (b) C4.5 as the base learner.}
(a) \\
\begin{tabular}{lr@{\hspace{0cm}}c@{\hspace{0cm}}r|r@{\hspace{0cm}}c@{\hspace{0cm}}r@{\hspace{0.1cm}}cr@{\hspace{0cm}}c@{\hspace{0cm}}r@{\hspace{0.1cm}}cr@{\hspace{0cm}}c@{\hspace{0cm}}r@{\hspace{0.1cm}}c}
\hline
Dataset & \multicolumn{3}{c}{RPND}& \multicolumn{4}{c}{NDBC} & \multicolumn{4}{c}{CBND} & \multicolumn{4}{c}{ND} \\
\hline
audiology 				& 75.36 & $\pm$ & 8.45 & 72.47 & $\pm$ & 8.80 &           & 68.55 & $\pm$ & 9.61 &           & 71.91 & $\pm$ & 9.85 &          \\
krkopt 					& 33.13 & $\pm$ & 0.97 & 33.23 & $\pm$ & 0.80 &           & 28.55 & $\pm$ & 1.50 & $\bullet$ & 28.70 & $\pm$ & 1.56 & $\bullet$\\
LED24 					& 72.85 & $\pm$ & 2.03 & 72.73 & $\pm$ & 2.06 &           & 67.11 & $\pm$ & 4.08 & $\bullet$ & 70.26 & $\pm$ & 3.28 & $\bullet$\\
letter 					& 67.70 & $\pm$ & 2.72 & 72.23 & $\pm$ & 0.93 &   $\circ$ & 47.98 & $\pm$ & 3.08 & $\bullet$ & 53.10 & $\pm$ & 4.36 & $\bullet$\\
mfeat-factors 			& 95.04 & $\pm$ & 1.99 & 96.62 & $\pm$ & 1.19 &   $\circ$ & 91.83 & $\pm$ & 2.20 & $\bullet$ & 93.08 & $\pm$ & 2.15 & $\bullet$\\
mfeat-fourier 			& 76.37 & $\pm$ & 3.22 & 75.17 & $\pm$ & 2.76 &           & 73.17 & $\pm$ & 3.34 & $\bullet$ & 74.00 & $\pm$ & 3.34 &          \\
mfeat-karhunen 			& 89.83 & $\pm$ & 2.32 & 90.83 & $\pm$ & 1.75 &           & 84.96 & $\pm$ & 3.75 & $\bullet$ & 86.53 & $\pm$ & 3.06 & $\bullet$\\
mfeat-morphological 		& 72.64 & $\pm$ & 3.25 & 70.45 & $\pm$ & 3.03 & $\bullet$ & 62.31 & $\pm$ & 7.79 & $\bullet$ & 66.40 & $\pm$ & 5.19 & $\bullet$\\
mfeat-pixel 				& 71.16 & $\pm$ & 9.98 & 88.67 & $\pm$ & 2.51 &   $\circ$ & 61.25 & $\pm$ & 9.25 &           & 47.44 & $\pm$ & 9.15 & $\bullet$\\
optdigits 				& 92.72 & $\pm$ & 2.06 & 92.00 & $\pm$ & 1.10 &           & 87.83 & $\pm$ & 3.01 & $\bullet$ & 90.95 & $\pm$ & 2.60 &          \\
page-blocks 				& 96.17 & $\pm$ & 0.75 & 95.77 & $\pm$ & 0.77 &           & 95.44 & $\pm$ & 0.84 & $\bullet$ & 95.61 & $\pm$ & 0.86 &          \\
pendigits 				& 90.20 & $\pm$ & 2.32 & 87.97 & $\pm$ & 0.96 & $\bullet$ & 82.23 & $\pm$ & 4.42 & $\bullet$ & 87.08 & $\pm$ & 4.22 &          \\
segment 					& 94.02 & $\pm$ & 2.40 & 88.76 & $\pm$ & 1.91 & $\bullet$ & 87.36 & $\pm$ & 4.16 & $\bullet$ & 89.11 & $\pm$ & 3.93 & $\bullet$\\
shuttle 					& 96.87 & $\pm$ & 0.46 & 96.86 & $\pm$ & 0.20 &           & 92.14 & $\pm$ & 6.86 &           & 91.72 & $\pm$ & 7.03 & $\bullet$\\
usps 					& 87.47 & $\pm$ & 1.47 & 87.64 & $\pm$ & 1.06 &           & 84.70 & $\pm$ & 2.26 & $\bullet$ & 85.83 & $\pm$ & 1.97 & $\bullet$\\
vowel 					& 81.80 & $\pm$ & 4.46 & 80.83 & $\pm$ & 4.10 &           & 47.86 & $\pm$ & 8.67 & $\bullet$ & 53.08 & $\pm$ & 8.98 & $\bullet$\\
yeast 					& 58.35 & $\pm$ & 3.89 & 59.00 & $\pm$ & 3.58 &           & 56.43 & $\pm$ & 4.20 &           & 55.91 & $\pm$ & 3.90 & $\bullet$\\
zoo 						& 90.41 & $\pm$ & 9.15 & 87.55 & $\pm$ & 9.32 &           & 88.88 & $\pm$ & 9.34 &           & 89.00 & $\pm$ & 8.65 &          \\
\hline
\end{tabular} \par}

{\centering 
(b) \\
\begin{tabular}{lr@{\hspace{0cm}}c@{\hspace{0cm}}r|r@{\hspace{0cm}}c@{\hspace{0cm}}r@{\hspace{0.1cm}}cr@{\hspace{0cm}}c@{\hspace{0cm}}r@{\hspace{0.1cm}}cr@{\hspace{0cm}}c@{\hspace{0cm}}r@{\hspace{0.1cm}}c}
\hline
Dataset & \multicolumn{3}{c}{RPND}& \multicolumn{4}{c}{NDBC} & \multicolumn{4}{c}{CBND} & \multicolumn{4}{c}{ND} \\
\hline
audiology 				& 76.86 & $\pm$ & 7.23 & 75.49 & $\pm$ & 7.29 &          & 74.45 & $\pm$ & 8.04 &           & 73.79 & $\pm$ & 7.62 &          \\
krkopt 					& 70.04 & $\pm$ & 2.45 & 69.33 & $\pm$ & 0.99 &          & 64.83 & $\pm$ & 1.78 & $\bullet$ & 65.13 & $\pm$ & 2.19 & $\bullet$\\
LED24 					& 72.68 & $\pm$ & 2.12 & 72.99 & $\pm$ & 1.72 &          & 72.07 & $\pm$ & 2.08 &           & 72.22 & $\pm$ & 2.05 &          \\
letter 					& 86.32 & $\pm$ & 0.85 & 86.50 & $\pm$ & 0.88 &          & 85.38 & $\pm$ & 0.88 & $\bullet$ & 86.03 & $\pm$ & 0.88 &          \\
mfeat-factors 			& 88.47 & $\pm$ & 2.59 & 88.77 & $\pm$ & 1.73 &          & 86.76 & $\pm$ & 2.43 &           & 87.47 & $\pm$ & 2.23 &          \\
mfeat-fourier 			& 74.46 & $\pm$ & 3.09 & 73.97 & $\pm$ & 2.90 &          & 72.63 & $\pm$ & 2.97 &           & 73.03 & $\pm$ & 3.29 &          \\
mfeat-karhunen 			& 82.04 & $\pm$ & 2.84 & 82.56 & $\pm$ & 2.66 &          & 80.11 & $\pm$ & 3.15 &           & 80.18 & $\pm$ & 3.28 &          \\
mfeat-morphological 		& 72.44 & $\pm$ & 2.73 & 72.27 & $\pm$ & 2.48 &          & 71.90 & $\pm$ & 2.40 &           & 71.85 & $\pm$ & 2.52 &          \\
mfeat-pixel 				& 81.83 & $\pm$ & 3.23 & 81.36 & $\pm$ & 2.79 &          & 77.13 & $\pm$ & 3.61 & $\bullet$ & 79.44 & $\pm$ & 3.91 &          \\
optdigits 				& 90.72 & $\pm$ & 1.43 & 90.76 & $\pm$ & 1.15 &          & 89.27 & $\pm$ & 1.52 & $\bullet$ & 89.93 & $\pm$ & 1.44 &          \\
page-blocks 				& 97.07 & $\pm$ & 0.72 & 97.05 & $\pm$ & 0.66 &          & 97.00 & $\pm$ & 0.67 &           & 97.05 & $\pm$ & 0.65 &          \\
pendigits 				& 95.92 & $\pm$ & 0.70 & 95.81 & $\pm$ & 0.62 &          & 95.60 & $\pm$ & 0.67 &           & 95.79 & $\pm$ & 0.68 &          \\
segment 					& 96.10 & $\pm$ & 1.38 & 96.59 & $\pm$ & 1.25 &          & 95.88 & $\pm$ & 1.49 &           & 95.88 & $\pm$ & 1.37 &          \\
shuttle 					& 99.97 & $\pm$ & 0.02 & 99.98 & $\pm$ & 0.02 &          & 99.97 & $\pm$ & 0.02 &           & 99.97 & $\pm$ & 0.03 &          \\
usps 					& 87.95 & $\pm$ & 1.18 & 89.44 & $\pm$ & 0.91 &  $\circ$ & 86.06 & $\pm$ & 1.52 & $\bullet$ & 86.48 & $\pm$ & 1.37 & $\bullet$\\
vowel 					& 79.04 & $\pm$ & 4.22 & 76.96 & $\pm$ & 4.45 &          & 76.07 & $\pm$ & 4.75 &           & 75.54 & $\pm$ & 4.87 &          \\
yeast 					& 57.22 & $\pm$ & 3.31 & 57.58 & $\pm$ & 3.69 &          & 56.18 & $\pm$ & 3.43 &           & 56.64 & $\pm$ & 3.36 &          \\
zoo 						& 91.63 & $\pm$ & 8.06 & 88.65 & $\pm$ & 8.30 &          & 90.72 & $\pm$ & 7.12 &           & 90.67 & $\pm$ & 8.72 &          \\
\hline
\end{tabular} \par}

\end{table}

Table~\ref{tab:single_classifier} shows the classification accuracy and standard deviations of each method when training a single nested dichotomy. When logistic regression is used as the base learner, compared to random methods (CBND and ND), we obtain a significant accuracy gain in most cases, and comparable accuracy in all others. When using C4.5 as the base learner, our method is preferable to random methods in some cases, with all other datasets showing a comparable accuracy.

In comparison to NDBC, gives similar accuracy overall, with three significantly better results, four significantly worse results, and the rest comparable over both base learners. It is to be expected that NDBC sometimes has better performance than our method when only a single nested dichotomy is built. This is because NDBC deterministically selects the class split that is likely to be the most easily separable. Our method attempts to produce an easily separable class subset selection from a pool of possible options, where each option is as likely as any other.

\subsection{Ensembles of Nested Dichotomies}
 
Ensembles of nested dichotomies typically outperform single nested dichotomies. The original method for creating an ensemble of nested dichotomies is a randomization approach, but it was later found that better performance can be obtained by bagging and boosting nested dichotomies~\cite{rodriguez2010forests}. For this reason, we consider three types of ensembles of nested dichotomies in our experiments: bagged, boosted with AdaBoost and boosted with MultiBoost (the latter two applied with resampling based on instance weights). We built ensembles of 10 nested dichotomies for these experiments.

\subsubsection*{Bagging.}

Table~\ref{tab:bagged_classifier} shows the results of using bagging to construct an ensemble of nested dichotomies for each method and for both base learners. When logistic regression is used as a base learner, our method outperforms all other methods in many cases. When C4.5 is used as a base learner, our method compares favourably with NDBC and achieves comparable accuracy to the random methods. Our method is better in a bagging scenario than NDBC because of the first problem highlighted in Section~\ref{sec:advantages},~\textit{i.e.}, using the furthest centroids to select a class  split results in a deterministic class split. Evidently, with bagged datasets, this method of class subset selection is too stable to be utilized effectively. Our method, on the other hand, is sufficiently unstable to be useful in a bagged ensemble. 

\begin{table}[t]
\scriptsize
{\centering 
\caption{\label{tab:bagged_classifier}Accuracy of an ensemble of 10 bagged nested dichotomies with (a) logistic regression and (b) C4.5 as the base learner.}
(a)\\
\begin{tabular}{lr@{\hspace{0cm}}c@{\hspace{0cm}}r|r@{\hspace{0cm}}c@{\hspace{0cm}}r@{\hspace{0.1cm}}cr@{\hspace{0cm}}c@{\hspace{0cm}}r@{\hspace{0.1cm}}cr@{\hspace{0cm}}c@{\hspace{0cm}}r@{\hspace{0.1cm}}c}
\hline
Dataset & \multicolumn{3}{c}{RPND}& \multicolumn{4}{c}{NDBC} & \multicolumn{4}{c}{CBND} & \multicolumn{4}{c}{ND} \\
\hline
audiology 				& 81.79 & $\pm$ & 7.56 & 81.25 & $\pm$ & 7.25 &           & 80.32 & $\pm$ & 7.69 &           & 82.35 & $\pm$ & 7.57 &          \\
krkopt 					& 33.77 & $\pm$ & 0.78 & 33.29 & $\pm$ & 0.77 & $\bullet$ & 31.73 & $\pm$ & 0.98 & $\bullet$ & 31.99 & $\pm$ & 0.94 & $\bullet$\\
LED24 					& 73.56 & $\pm$ & 1.90 & 73.42 & $\pm$ & 2.01 &           & 73.50 & $\pm$ & 1.94 &           & 73.49 & $\pm$ & 1.85 &          \\
letter 					& 78.65 & $\pm$ & 0.94 & 76.16 & $\pm$ & 0.96 & $\bullet$ & 73.76 & $\pm$ & 1.24 & $\bullet$ & 74.51 & $\pm$ & 1.27 & $\bullet$\\
mfeat-factors 			& 98.11 & $\pm$ & 1.02 & 97.39 & $\pm$ & 1.10 & $\bullet$ & 97.72 & $\pm$ & 1.09 &           & 97.94 & $\pm$ & 1.01 &          \\
mfeat-fourier 			& 83.08 & $\pm$ & 2.18 & 80.03 & $\pm$ & 2.25 & $\bullet$ & 82.16 & $\pm$ & 2.66 &           & 82.14 & $\pm$ & 2.39 &          \\
mfeat-karhunen 			& 95.66 & $\pm$ & 1.54 & 93.67 & $\pm$ & 1.75 & $\bullet$ & 94.88 & $\pm$ & 1.56 &           & 94.89 & $\pm$ & 1.57 &          \\
mfeat-morphological 		& 73.71 & $\pm$ & 2.79 & 72.33 & $\pm$ & 2.87 &           & 73.19 & $\pm$ & 2.94 &           & 73.55 & $\pm$ & 2.45 &          \\
mfeat-pixel 				& 94.70 & $\pm$ & 1.95 & 93.15 & $\pm$ & 1.49 & $\bullet$ & 90.96 & $\pm$ & 2.51 & $\bullet$ & 83.65 & $\pm$ & 4.01 & $\bullet$\\
optdigits 				& 97.15 & $\pm$ & 0.68 & 93.56 & $\pm$ & 0.93 & $\bullet$ & 96.50 & $\pm$ & 0.83 & $\bullet$ & 96.83 & $\pm$ & 0.68 &          \\
page-blocks 				& 96.46 & $\pm$ & 0.68 & 96.14 & $\pm$ & 0.66 & $\bullet$ & 95.92 & $\pm$ & 0.72 & $\bullet$ & 96.11 & $\pm$ & 0.68 & $\bullet$\\
pendigits 				& 95.93 & $\pm$ & 0.80 & 88.90 & $\pm$ & 1.08 & $\bullet$ & 94.61 & $\pm$ & 1.00 & $\bullet$ & 95.12 & $\pm$ & 0.88 & $\bullet$\\
segment 					& 95.37 & $\pm$ & 1.61 & 89.26 & $\pm$ & 1.95 & $\bullet$ & 94.03 & $\pm$ & 1.96 & $\bullet$ & 94.15 & $\pm$ & 1.73 & $\bullet$\\
shuttle 					& 96.74 & $\pm$ & 0.24 & 96.86 & $\pm$ & 0.21 & 		     & 94.94 & $\pm$ & 1.52 & $\bullet$ & 94.86 & $\pm$ & 1.39 & $\bullet$\\
usps 					& 93.83 & $\pm$ & 0.69 & 92.02 & $\pm$ & 0.91 & $\bullet$ & 93.59 & $\pm$ & 0.70 &           & 93.32 & $\pm$ & 0.73 & $\bullet$\\
vowel 					& 89.76 & $\pm$ & 3.04 & 85.72 & $\pm$ & 3.49 & $\bullet$ & 77.52 & $\pm$ & 4.90 & $\bullet$ & 78.30 & $\pm$ & 4.61 & $\bullet$\\
yeast 					& 58.86 & $\pm$ & 3.85 & 59.18 & $\pm$ & 3.84 &           & 58.91 & $\pm$ & 3.64 &           & 58.92 & $\pm$ & 3.62 &          \\
zoo 						& 94.87 & $\pm$ & 6.03 & 91.62 & $\pm$ & 8.33 &           & 93.36 & $\pm$ & 7.16 &           & 93.20 & $\pm$ & 7.37 &          \\
\hline
\end{tabular} \par}

{\centering
(b) \\
\begin{tabular}{lr@{\hspace{0cm}}c@{\hspace{0cm}}r|r@{\hspace{0cm}}c@{\hspace{0cm}}r@{\hspace{0.1cm}}cr@{\hspace{0cm}}c@{\hspace{0cm}}r@{\hspace{0.1cm}}cr@{\hspace{0cm}}c@{\hspace{0cm}}r@{\hspace{0.1cm}}c}
\hline
Dataset & \multicolumn{3}{c}{RPND}& \multicolumn{4}{c}{NDBC} & \multicolumn{4}{c}{CBND} & \multicolumn{4}{c}{ND} \\
\hline								
audiology 				& 79.76 & $\pm$ & 7.32 & 80.33 & $\pm$ & 6.11 &           & 80.65 & $\pm$ & 7.29 &           & 79.30 & $\pm$ & 7.30 &          \\
krkopt 					& 75.70 & $\pm$ & 0.95 & 73.93 & $\pm$ & 0.90 & $\bullet$ & 74.20 & $\pm$ & 1.00 & $\bullet$ & 74.82 & $\pm$ & 1.00 & $\bullet$\\
LED24 					& 73.22 & $\pm$ & 1.92 & 73.12 & $\pm$ & 1.82 &           & 73.10 & $\pm$ & 1.90 &           & 73.23 & $\pm$ & 1.92 &          \\
letter 					& 93.81 & $\pm$ & 0.55 & 92.73 & $\pm$ & 0.66 & $\bullet$ & 93.92 & $\pm$ & 0.50 &           & 94.07 & $\pm$ & 0.49 &          \\
mfeat-factors 			& 95.27 & $\pm$ & 1.58 & 93.37 & $\pm$ & 1.76 & $\bullet$ & 95.80 & $\pm$ & 1.40 &           & 95.44 & $\pm$ & 1.52 &          \\
mfeat-fourier 			& 81.36 & $\pm$ & 2.81 & 78.79 & $\pm$ & 2.64 & $\bullet$ & 81.30 & $\pm$ & 2.83 &           & 80.94 & $\pm$ & 2.76 &          \\
mfeat-karhunen 			& 92.83 & $\pm$ & 1.96 & 90.27 & $\pm$ & 2.11 & $\bullet$ & 92.98 & $\pm$ & 1.42 &           & 93.13 & $\pm$ & 1.67 &          \\
mfeat-morphological 		& 73.38 & $\pm$ & 2.61 & 72.78 & $\pm$ & 2.72 &           & 73.07 & $\pm$ & 2.83 &           & 73.37 & $\pm$ & 2.62 &          \\
mfeat-pixel 				& 92.56 & $\pm$ & 1.91 & 87.01 & $\pm$ & 2.47 & $\bullet$ & 92.24 & $\pm$ & 1.82 &           & 92.65 & $\pm$ & 1.79 &          \\
optdigits 				& 97.09 & $\pm$ & 0.70 & 95.34 & $\pm$ & 0.90 & $\bullet$ & 97.04 & $\pm$ & 0.72 &           & 97.00 & $\pm$ & 0.72 &          \\
page-blocks 				& 97.41 & $\pm$ & 0.64 & 97.29 & $\pm$ & 0.62 &           & 97.39 & $\pm$ & 0.59 &           & 97.36 & $\pm$ & 0.63 &          \\
pendigits 				& 98.53 & $\pm$ & 0.40 & 97.67 & $\pm$ & 0.46 & $\bullet$ & 98.68 & $\pm$ & 0.35 &           & 98.64 & $\pm$ & 0.38 &          \\
segment 					& 97.45 & $\pm$ & 1.09 & 97.52 & $\pm$ & 1.11 &           & 97.54 & $\pm$ & 1.14 &           & 97.53 & $\pm$ & 0.88 &          \\
shuttle 					& 99.98 & $\pm$ & 0.02 & 99.97 & $\pm$ & 0.02 &           & 99.98 & $\pm$ & 0.02 &           & 99.98 & $\pm$ & 0.02 &          \\
usps 					& 94.63 & $\pm$ & 0.59 & 93.85 &	$\pm$ &	0.72	 & $\bullet$ & 94.52 & $\pm$ & 0.59 &           & 94.61 & $\pm$ & 0.70 &          \\
vowel 					& 87.69 & $\pm$ & 3.52 & 85.82 & $\pm$ & 3.73 &           & 89.15 & $\pm$ & 3.46 &           & 88.26 & $\pm$ & 3.25 &          \\
yeast 					& 59.86 & $\pm$ & 3.29 & 59.55 & $\pm$ & 3.38 &           & 59.93 & $\pm$ & 3.54 &           & 59.72 & $\pm$ & 3.79 &          \\
zoo 						& 93.81 & $\pm$ & 7.17 & 91.70 & $\pm$ & 7.77 &           & 93.57 & $\pm$ & 6.81 &           & 94.36 & $\pm$ & 6.17 &          \\
\hline

\end{tabular} \scriptsize \par}

\end{table}
\subsubsection*{AdaBoost.}

Table~\ref{tab:adaboost_classifier} shows the results of using AdaBoost to build an ensemble of nested dichotomies for each method and for both base learners. When comparing with the random methods, we observe a similar result to the bagged ensembles. When using logistic regression, we see a significant improvement in accuracy in many cases, and when C4.5 is used, we typically see comparable results, with a small number of significant accuracy gains. When comparing with NDBC, we see a small improvement for the vast majority of datasets, but these differences are almost never individually significant. In one instance (krkopt with C4.5 as the base learner), we achieve a significant accuracy gain using our method. 

\begin{table}[t]
\scriptsize
{\centering 
\caption{\label{tab:adaboost_classifier}Accuracy of an ensemble of 10 nested dichotomies boosted with AdaBoost with (a) logistic regression and (b) C4.5 as the base learner.}
(a) \\
\begin{tabular}{lr@{\hspace{0cm}}c@{\hspace{0cm}}r|r@{\hspace{0cm}}c@{\hspace{0cm}}r@{\hspace{0.1cm}}cr@{\hspace{0cm}}c@{\hspace{0cm}}r@{\hspace{0.1cm}}cr@{\hspace{0cm}}c@{\hspace{0cm}}r@{\hspace{0.1cm}}c}
\hline
Dataset & \multicolumn{3}{c}{RPND}& \multicolumn{4}{c}{NDBC} & \multicolumn{4}{c}{CBND} & \multicolumn{4}{c}{ND} \\
\hline
audiology 				& 81.42 & $\pm$ & 7.38 & 80.31 & $\pm$ & 6.92 &          & 79.87 & $\pm$ &  7.49 &           & 80.78 & $\pm$ &  7.50 &          \\
krkopt 					& 32.99 & $\pm$ & 1.01 & 32.81 & $\pm$ & 0.77 &          & 28.24 & $\pm$ &  1.47 & $\bullet$ & 28.66 & $\pm$ &  1.44 & $\bullet$\\
LED24 					& 72.41 & $\pm$ & 2.16 & 72.93 & $\pm$ & 1.99 &          & 69.17 & $\pm$ &  2.77 & $\bullet$ & 70.44 & $\pm$ &  2.72 & $\bullet$\\
letter 					& 71.39 & $\pm$ & 2.50 & 71.44 & $\pm$ & 1.49 &          & 47.42 & $\pm$ &  3.29 & $\bullet$ & 55.16 & $\pm$ &  5.35 & $\bullet$\\
mfeat-factors 			& 97.71 & $\pm$ & 1.09 & 97.66 & $\pm$ & 0.99 &          & 97.11 & $\pm$ &  1.25 &           & 97.52 & $\pm$ &  1.17 &          \\
mfeat-fourier 			& 81.01 & $\pm$ & 2.28 & 79.96 & $\pm$ & 2.52 &          & 80.12 & $\pm$ &  2.43 &           & 80.13 & $\pm$ &  2.64 &          \\
mfeat-karhunen 			& 94.93 & $\pm$ & 1.50 & 94.42 & $\pm$ & 1.61 &          & 93.76 & $\pm$ &  1.54 & $\bullet$ & 94.01 & $\pm$ &  1.54 &          \\
mfeat-morphological 		& 72.81 & $\pm$ & 2.82 & 71.02 & $\pm$ & 3.10 &          & 66.73 & $\pm$ &  6.80 & $\bullet$ & 69.38 & $\pm$ &  5.53 &          \\
mfeat-pixel 				& 94.15 & $\pm$ & 1.81 & 93.87 & $\pm$ & 1.59 &          & 91.16 & $\pm$ &  2.39 & $\bullet$ & 86.21 & $\pm$ &  3.48 & $\bullet$\\
nursery 					& 92.51 & $\pm$ & 0.70 & 92.52 & $\pm$ & 0.70 &          & 92.29 & $\pm$ &  0.74 &           & 92.38 & $\pm$ &  0.69 &          \\
optdigits 				& 97.01 & $\pm$ & 0.69 & 96.84 & $\pm$ & 0.77 &          & 96.26 & $\pm$ &  0.74 & $\bullet$ & 96.37 & $\pm$ &  0.86 & $\bullet$\\
page-blocks 				& 96.09 & $\pm$ & 0.80 & 95.93 & $\pm$ & 0.75 &          & 95.43 & $\pm$ &  0.84 &           & 95.77 & $\pm$ &  0.90 &          \\
pendigits 				& 94.94 & $\pm$ & 0.93 & 94.83 & $\pm$ & 0.77 &          & 93.86 & $\pm$ &  1.30 &           & 93.67 & $\pm$ &  1.03 & $\bullet$\\
segment 					& 94.94 & $\pm$ & 1.40 & 94.66 & $\pm$ & 1.48 &          & 93.88 & $\pm$ &  1.93 &           & 93.82 & $\pm$ &  1.84 &          \\
shuttle 					& 96.83 & $\pm$ & 0.45 & 96.86 & $\pm$ & 0.26 &          & 96.51 & $\pm$ &  1.57 &           & 96.40 & $\pm$ &  2.18 &          \\
usps 					& 92.03 & $\pm$ & 0.88 & 91.83 & $\pm$ & 0.86 &          & 91.91 & $\pm$ &  0.91 &           & 91.66 & $\pm$ &  0.85 &          \\
vowel 					& 90.59 & $\pm$ & 3.11 & 89.74 & $\pm$ & 3.10 &          & 48.45 & $\pm$ & 10.68 & $\bullet$ & 58.93 & $\pm$ & 11.42 & $\bullet$\\
yeast 					& 57.97 & $\pm$ & 3.78 & 58.39 & $\pm$ & 3.62 &          & 56.90 & $\pm$ &  4.05 &           & 56.56 & $\pm$ &  3.66 &          \\
zoo 						& 94.95 & $\pm$ & 6.40 & 94.96 & $\pm$ & 6.33 &          & 94.38 & $\pm$ &  7.44 &           & 94.77 & $\pm$ &  6.19 &          \\
\hline
\end{tabular} \par}

{\centering 
(b) \\
\begin{tabular}{lr@{\hspace{0cm}}c@{\hspace{0cm}}r|r@{\hspace{0cm}}c@{\hspace{0cm}}r@{\hspace{0.1cm}}cr@{\hspace{0cm}}c@{\hspace{0cm}}r@{\hspace{0.1cm}}cr@{\hspace{0cm}}c@{\hspace{0cm}}r@{\hspace{0.1cm}}c}
\hline
Dataset & \multicolumn{3}{c}{RPND}& \multicolumn{4}{c}{NDBC} & \multicolumn{4}{c}{CBND} & \multicolumn{4}{c}{ND} \\
\hline
audiology 				& 83.64 & $\pm$ & 7.37 & 83.29 & $\pm$ & 6.68 &			 & 82.63 & $\pm$ & 6.87 &           & 82.58 & $\pm$ & 7.36 &          \\
krkopt 					& 81.01 & $\pm$ & 0.78 & 79.37 & $\pm$ & 0.80 & $\bullet$ & 77.25 & $\pm$ & 0.95 & $\bullet$ & 78.36 & $\pm$ & 1.04 & $\bullet$\\
LED24 					& 69.59 & $\pm$ & 2.13 & 69.49 & $\pm$ & 2.11 &			 & 69.04 & $\pm$ & 1.95 &           & 69.42 & $\pm$ & 1.78 &          \\
letter 					& 94.58 & $\pm$ & 0.49 & 94.37 & $\pm$ & 0.48 &			 & 94.30 & $\pm$ & 0.49 &           & 94.60 & $\pm$ & 0.55 &          \\
mfeat-factors 			& 95.75 & $\pm$ & 1.36 & 95.31 & $\pm$ & 1.48 &			 & 95.49 & $\pm$ & 1.38 &           & 95.62 & $\pm$ & 1.37 &          \\
mfeat-fourier 			& 80.43 & $\pm$ & 2.74 & 79.54 & $\pm$ & 2.60 & 			 & 80.12 & $\pm$ & 2.49 &           & 80.74 & $\pm$ & 2.47 &          \\
mfeat-karhunen 			& 93.20 & $\pm$ & 1.80 & 92.67 & $\pm$ & 1.83 &			 & 92.96 & $\pm$ & 1.76 &           & 92.85 & $\pm$ & 1.84 &          \\
mfeat-morphological 		& 70.48 & $\pm$ & 3.10 & 70.45 & $\pm$ & 3.19 &			 & 70.13 & $\pm$ & 2.84 &           & 70.50 & $\pm$ & 2.45 &          \\
mfeat-pixel 				& 93.76 & $\pm$ & 1.53 & 93.27 & $\pm$ & 1.80 &			 & 92.48 & $\pm$ & 1.80 & $\bullet$ & 93.01 & $\pm$ & 1.83 &          \\
optdigits 				& 97.31 & $\pm$ & 0.72 & 97.23 & $\pm$ & 0.70 &			 & 97.25 & $\pm$ & 0.68 &           & 97.20 & $\pm$ & 0.70 &          \\
page-blocks 				& 97.05 & $\pm$ & 0.62 & 97.05 & $\pm$ & 0.66 &			 & 97.11 & $\pm$ & 0.64 &           & 97.11 & $\pm$ & 0.66 &          \\
pendigits 				& 98.95 & $\pm$ & 0.30 & 98.89 & $\pm$ & 0.33 &			 & 98.91 & $\pm$ & 0.30 &           & 98.93 & $\pm$ & 0.28 &          \\
segment 					& 98.23 & $\pm$ & 0.84 & 98.24 & $\pm$ & 0.84 &			 & 98.09 & $\pm$ & 0.86 &           & 98.09 & $\pm$ & 0.94 &          \\
shuttle 					& 99.99 & $\pm$ & 0.01 & 99.99 & $\pm$ & 0.01 &			 & 99.99 & $\pm$ & 0.01 &           & 99.99 & $\pm$ & 0.01 &          \\
usps 					& 94.85 & $\pm$ & 0.64 & 94.86 & $\pm$ & 0.64 &			 & 94.41 & $\pm$ & 0.72 &           & 94.59 & $\pm$ & 0.66 &          \\
vowel 					& 91.95 & $\pm$ & 2.71 & 90.73 & $\pm$ & 3.00 &			 & 91.28 & $\pm$ & 2.82 &           & 91.30 & $\pm$ & 2.78 &          \\
yeast 					& 57.39 & $\pm$ & 3.76 & 57.42 & $\pm$ & 4.02 &			 & 56.93 & $\pm$ & 3.27 &           & 57.25 & $\pm$ & 4.19 &          \\
zoo 						& 95.45 & $\pm$ & 6.19 & 95.53 & $\pm$ & 6.39 &			 & 95.15 & $\pm$ & 6.21 &           & 95.36 & $\pm$ & 6.13 &          \\
\hline
\end{tabular}  \par}

\end{table}

\subsubsection*{MultiBoost.}

Table~\ref{tab:multiboost_classifier} shows the results of using MultiBoost to build an ensemble of nested dichotomies for each method and for both base learners. Compared to the random methods, again we see similar results to the other ensemble methods -- using logistic regression as the base learner results in many significant improvements, and using C4.5 as the base learner typically produces comparable results, with few significant improvements. In comparison to NDBC, we see many small (although statistically insignificant) improvements across both base learners, with some significant gains in accuracy on some datasets. 

\begin{table}[t]
\caption{\label{tab:multiboost_classifier}Accuracy of an ensemble of 10 nested dichotomies boosted with MultiBoost with (a) logistic regression and (b) C4.5 as the base learner.}

\scriptsize
{\centering 
(a) \\
\begin{tabular}{lr@{\hspace{0cm}}c@{\hspace{0cm}}r|r@{\hspace{0cm}}c@{\hspace{0cm}}r@{\hspace{0.1cm}}cr@{\hspace{0cm}}c@{\hspace{0cm}}r@{\hspace{0.1cm}}cr@{\hspace{0cm}}c@{\hspace{0cm}}r@{\hspace{0.1cm}}c}
\hline
Dataset & \multicolumn{3}{c}{RPND}& \multicolumn{4}{c}{NDBC} & \multicolumn{4}{c}{CBND} & \multicolumn{4}{c}{ND} \\
\hline
audiology 			& 80.55 & $\pm$ & 7.80 & 80.05 & $\pm$ & 7.20 &           & 78.90 & $\pm$ &  7.51 &           & 79.53 & $\pm$ &  7.73 &          \\
krkopt 				& 32.99 & $\pm$ & 1.01 & 32.81 & $\pm$ & 0.77 &           & 28.24 & $\pm$ &  1.47 & $\bullet$ & 28.66 & $\pm$ &  1.44 & $\bullet$\\
LED24 				& 73.38 & $\pm$ & 1.81 & 73.31 & $\pm$ & 2.15 &           & 72.01 & $\pm$ &  2.67 &           & 72.75 & $\pm$ &  2.38 &          \\
letter 				& 77.29 & $\pm$ & 1.83 & 75.36 & $\pm$ & 1.03 & $\bullet$ & 47.42 & $\pm$ &  3.29 & $\bullet$ & 55.85 & $\pm$ &  6.25 & $\bullet$\\
mfeat-factors		& 97.82 & $\pm$ & 1.16 & 97.70 & $\pm$ & 1.09 &           & 97.40 & $\pm$ &  1.31 &           & 97.53 & $\pm$ &  1.17 &          \\
mfeat-fourier 		& 82.12 & $\pm$ & 2.28 & 80.22 & $\pm$ & 2.28 & $\bullet$ & 80.22 & $\pm$ &  2.35 & $\bullet$ & 80.72 & $\pm$ &  2.44 &          \\
mfeat-karhunen 		& 95.22 & $\pm$ & 1.59 & 94.70 & $\pm$ & 1.57 &           & 93.94 & $\pm$ &  1.62 & $\bullet$ & 94.17 & $\pm$ &  1.71 &          \\
mfeat-morphological 	& 73.63 & $\pm$ & 2.80 & 72.33 & $\pm$ & 2.64 &           & 67.52 & $\pm$ &  7.04 & $\bullet$ & 70.40 & $\pm$ &  5.74 &          \\
mfeat-pixel 			& 94.37 & $\pm$ & 1.48 & 94.16 & $\pm$ & 1.30 &           & 91.89 & $\pm$ &  2.71 & $\bullet$ & 86.37 & $\pm$ &  4.74 & $\bullet$\\
optdigits 			& 97.03 & $\pm$ & 0.57 & 96.10 & $\pm$ & 0.79 & $\bullet$ & 96.26 & $\pm$ &  0.78 & $\bullet$ & 96.47 & $\pm$ &  0.83 & $\bullet$\\
page-blocks 			& 96.39 & $\pm$ & 0.69 & 96.10 & $\pm$ & 0.72 &           & 96.01 & $\pm$ &  0.68 & $\bullet$ & 96.19 & $\pm$ &  0.74 &          \\
pendigits 			& 96.02 & $\pm$ & 0.73 & 94.27 & $\pm$ & 1.32 & $\bullet$ & 94.17 & $\pm$ &  1.05 & $\bullet$ & 94.76 & $\pm$ &  0.92 & $\bullet$\\
segment 				& 95.56 & $\pm$ & 1.40 & 94.11 & $\pm$ & 1.92 & $\bullet$ & 94.12 & $\pm$ &  1.93 & $\bullet$ & 94.35 & $\pm$ &  1.63 & $\bullet$\\
shuttle 				& 96.89 & $\pm$ & 0.27 & 96.87 & $\pm$ & 0.24 &           & 96.63 & $\pm$ &  1.53 &           & 96.65 & $\pm$ &  1.59 &          \\
usps 				& 93.12 & $\pm$ & 0.78 & 92.45 & $\pm$ & 0.84 & $\bullet$ & 92.62 & $\pm$ &  0.83 &           & 92.57 & $\pm$ &  0.84 &          \\
vowel 				& 89.53 & $\pm$ & 3.15 & 87.52 & $\pm$ & 3.03 &           & 48.92 & $\pm$ & 11.26 & $\bullet$ & 60.91 & $\pm$ & 12.38 & $\bullet$\\
yeast 				& 58.28 & $\pm$ & 4.19 & 58.60 & $\pm$ & 3.93 &           & 57.13 & $\pm$ &  4.03 &           & 57.03 & $\pm$ &  3.88 &          \\
zoo 					& 94.97 & $\pm$ & 6.49 & 94.65 & $\pm$ & 6.79 &           & 94.46 & $\pm$ &  7.35 &           & 94.07 & $\pm$ &  7.02 &          \\
\hline
\end{tabular} \scriptsize \par}

\scriptsize
{\centering 
(b)\\
\begin{tabular}{lr@{\hspace{0cm}}c@{\hspace{0cm}}r|r@{\hspace{0cm}}c@{\hspace{0cm}}r@{\hspace{0.1cm}}cr@{\hspace{0cm}}c@{\hspace{0cm}}r@{\hspace{0.1cm}}cr@{\hspace{0cm}}c@{\hspace{0cm}}r@{\hspace{0.1cm}}c}
\hline
Dataset & \multicolumn{3}{c}{RPND}& \multicolumn{4}{c}{NDBC} & \multicolumn{4}{c}{CBND} & \multicolumn{4}{c}{ND} \\
\hline
audiology 			& 81.32 & $\pm$ & 7.06 & 82.14 & $\pm$ & 7.39 &           & 81.25 & $\pm$ & 7.48 &           & 80.32 & $\pm$ & 7.37 &          \\
krkopt 				& 76.22 & $\pm$ & 0.80 & 75.05 & $\pm$ & 0.84 & $\bullet$ & 73.54 & $\pm$ & 1.03 & $\bullet$ & 74.58 & $\pm$ & 1.14 & $\bullet$\\
LED24 				& 72.27 & $\pm$ & 2.00 & 71.90 & $\pm$ & 1.99 &           & 71.78 & $\pm$ & 1.89 &           & 71.96 & $\pm$ & 1.99 &          \\
letter 				& 93.98 & $\pm$ & 0.47 & 93.65 & $\pm$ & 0.53 &           & 93.78 & $\pm$ & 0.55 &           & 93.98 & $\pm$ & 0.46 &          \\
mfeat-factors 		& 95.63 & $\pm$ & 1.33 & 94.82 & $\pm$ & 1.45 &           & 95.32 & $\pm$ & 1.46 &           & 95.14 & $\pm$ & 1.48 &          \\
mfeat-fourier 		& 80.46 & $\pm$ & 2.40 & 79.54 & $\pm$ & 2.36 &           & 80.36 & $\pm$ & 2.57 &           & 80.68 & $\pm$ & 3.00 &          \\
mfeat-karhunen 		& 92.88 & $\pm$ & 1.95 & 91.82 & $\pm$ & 1.91 &           & 92.16 & $\pm$ & 2.03 &           & 92.64 & $\pm$ & 1.81 &          \\
mfeat-morphological 	& 71.30 & $\pm$ & 2.75 & 71.26 & $\pm$ & 2.85 &           & 71.32 & $\pm$ & 3.11 &           & 71.75 & $\pm$ & 2.84 &          \\
mfeat-pixel 			& 93.10 & $\pm$ & 1.71 & 91.15 & $\pm$ & 1.86 & $\bullet$ & 91.75 & $\pm$ & 1.67 & $\bullet$ & 92.40 & $\pm$ & 1.90 &        	\\
optdigits 			& 97.00 & $\pm$ & 0.70 & 96.80 & $\pm$ & 0.75 &           & 96.91 & $\pm$ & 0.73 &           & 97.00 & $\pm$ & 0.69 &          \\
page-blocks 			& 97.33 & $\pm$ & 0.65 & 97.24 & $\pm$ & 0.63 &           & 97.34 & $\pm$ & 0.64 &           & 97.29 & $\pm$ & 0.66 &          \\
pendigits 			& 98.78 & $\pm$ & 0.35 & 98.69 & $\pm$ & 0.35 &           & 98.78 & $\pm$ & 0.33 &           & 98.75 & $\pm$ & 0.28 &          \\
segment 				& 97.90 & $\pm$ & 0.93 & 98.06 & $\pm$ & 0.94 &           & 97.79 & $\pm$ & 0.95 &           & 97.88 & $\pm$ & 0.99 &          \\
shuttle 				& 99.99 & $\pm$ & 0.02 & 99.99 & $\pm$ & 0.02 &           & 99.99 & $\pm$ & 0.02 &           & 99.99 & $\pm$ & 0.01 &          \\
usps 				& 94.67 & $\pm$ & 0.65 & 94.48 & $\pm$ & 0.64 &           & 94.25 & $\pm$ & 0.58 &           & 94.33 & $\pm$ & 0.71 &          \\
vowel 				& 88.60 & $\pm$ & 3.40 & 88.33 & $\pm$ & 3.61 &           & 88.79 & $\pm$ & 3.18 &           & 88.34 & $\pm$ & 3.56 &          \\
yeast 				& 58.91 & $\pm$ & 3.58 & 58.91 & $\pm$ & 3.56 &           & 58.53 & $\pm$ & 3.63 &           & 58.35 & $\pm$ & 3.92 &          \\
zoo 					& 95.09 & $\pm$ & 6.73 & 94.17 & $\pm$ & 7.34 &           & 94.26 & $\pm$ & 6.48 &           & 95.66 & $\pm$ & 6.11 &          \\
\hline
\end{tabular} \scriptsize \par}
\end{table}

\subsection{\label{sec:train_time}Training Time}
\begin{figure*}[t]
	\centering
	\begin{subfigure}{6cm}
		\includegraphics[scale=0.65]{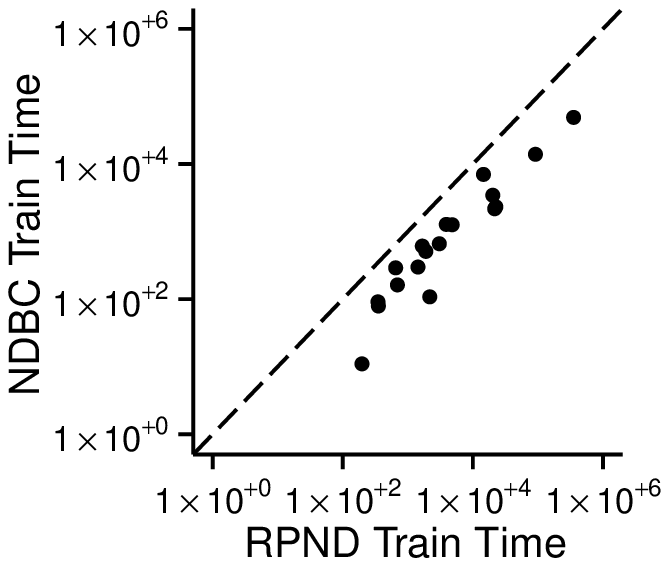}
		\caption{\label{fig:logistic_classes}Logistic regression}
	\end{subfigure}
	\begin{subfigure}{6cm}
		\includegraphics[scale=0.65]{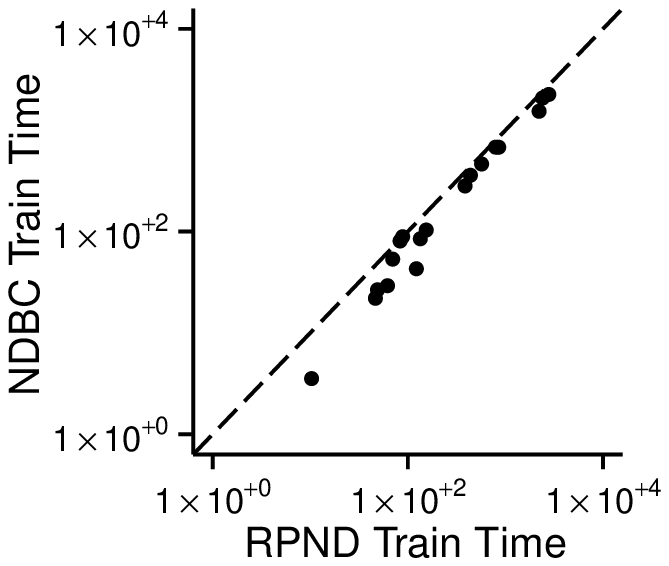}
		\caption{\label{fig:j48_classes}C4.5}
	\end{subfigure}	
	
	\caption{\label{fig:train_time}Log-log plots of the training time for a single RPND and a single NDBC, for both base learners.}
\end{figure*}

Fig.~\ref{fig:train_time} shows the training time in milliseconds for training a single RPND and a single NDBC, with logistic regression and C4.5 as the base learners for each of the datasets used in this evaluation. As can be seen from the plots, there is a computational cost for building an RPND over an NDBC, which is to be expected as there is an additional classifier trained and tested at each split node of the tree. The gradient of both plots is approximately one, which indicates that our method does not add additional computational complexity to the problem. The runtime is comparatively worse for logistic regression than for C4.5.

\subsection{\label{sec:case_study}Case Study: CIFAR-10}
\begin{figure*}[t]
	\centering
	\begin{subfigure}{\textwidth}
	\centering
	\resizebox{0.95\textwidth}{!}{
	\begin{tikzpicture}
		\tikzstyle{level 1}=[sibling distance = 90mm]
		\tikzstyle{level 2}=[sibling distance = 45mm]
		\tikzstyle{level 3}=[sibling distance = 25mm]
		\usetikzlibrary{shapes}
		\node[ellipse,draw,align=center](z){all classes}
		  child
		  {
		  	node[ellipse,draw,align=center]{truck, ship, \\ plane, car}
		  	child
		  	{
		  		node[ellipse,draw,align=center]{plane, ship}
		  		child { node[ellipse,draw,align=center]{plane} }
		  		child { node[ellipse,draw,align=center]{ship} }
		  	}
		  	child
		  	{
		  		node[ellipse,draw,align=center]{truck, car}
		  		child { node[ellipse,draw,align=center]{truck} }
		  		child { node[ellipse,draw,align=center]{car} }
		  	}
		  }
		  child
		  {
	    		node[ellipse,draw,align=center]{dog, horse, bird, \\ frog, deer, cat} 
	    		child
	    		{
	    			node[ellipse,draw,align=center] {cat, dog,\\ frog}
	    			child 
	    			{ 
	    				node[ellipse,draw,align=center]{cat, frog} 
	    				child { node[ellipse,draw,align=center]{cat} }
		  			child { node[ellipse,draw,align=center]{frog} }
	    			}
		  		child { node[ellipse,draw,align=center]{dog} }
	    		} 
	    		child
	    		{
	    			node[ellipse,draw,align=center] {deer, bird,\\ horse}
	    			child 
	    			{ 
	    				node[ellipse,draw,align=center]{deer, bird} 
	    				child { node[ellipse,draw,align=center]{deer} }
		  			child { node[ellipse,draw,align=center]{bird} }
	    			}
		  		child { node[ellipse,draw,align=center]{horse} }
	    		}
	    	  };
	\end{tikzpicture}
	}
	\caption{}
	\end{subfigure}
	\begin{subfigure}{\textwidth}
	\centering
	\resizebox{0.95\textwidth}{!}{
		\begin{tikzpicture}
		\tikzstyle{level 1}=[sibling distance = 60mm]
		\tikzstyle{level 2}=[sibling distance = 40mm]
		\tikzstyle{level 3}=[sibling distance = 45mm]
		\tikzstyle{level 4}=[sibling distance = 25mm]
		\usetikzlibrary{shapes}
		\node[ellipse,draw,align=center](z){all classes}
		  child
		  {
		  	node[ellipse,draw,align=center]{plane, ship,\\ truck}
		  	child 
	    		{ 
	    			node[ellipse,draw,align=center]{plane, ship} 
	    			child { node[ellipse,draw,align=center]{plane} }
		  		child { node[ellipse,draw,align=center]{ship} }
	    		}
		  	child { node[ellipse,draw,align=center]{truck} }
		  }
		  child
		  {
	    		node[ellipse,draw,align=center]{dog, horse, bird, \\ frog, deer, cat, car} 
	    		child
	    		{
	    			node[ellipse,draw,align=center] {car}
	    		} 
	    		child
	    		{
	    			node[ellipse,draw,align=center] {dog, horse, bird, \\ frog, deer, cat}
	    			child
	    			{
	    				node[ellipse,draw,align=center] {frog, deer,\\ dog}
		    			child 
		    			{ 
		    				node[ellipse,draw,align=center]{deer, frog} 
		    				child { node[ellipse,draw,align=center]{deer} }
			  			child { node[ellipse,draw,align=center]{frog} }
		    			}
			  		child { node[ellipse,draw,align=center]{dog} }
		    		} 
		    		child
		    		{
		    			node[ellipse,draw,align=center] {cat, bird,\\ horse}
		    			child 
		    			{ 
		    				node[ellipse,draw,align=center]{horse, bird} 
		    				child 
		    				{ 
		    					node[ellipse,draw,align=center]{horse} 
		    				}
				  		child 
				  		{ 
				  			node[ellipse,draw,align=center]{bird} 
				  		}
	    				}
			  		child 
			  		{ 
			  			node[ellipse,draw,align=center]{cat}
			  			child[missing]{}
			  			child[missing]{}
			  		}
	    			}
	    		}
	    	  };
	\end{tikzpicture}
	}
	\caption{}
	\end{subfigure}
	\caption{\label{fig:cnn_trees} Nested dichotomies trained on CIFAR-10, with (a) random-pair selection, and (b) centroid-based selection.}
\end{figure*}
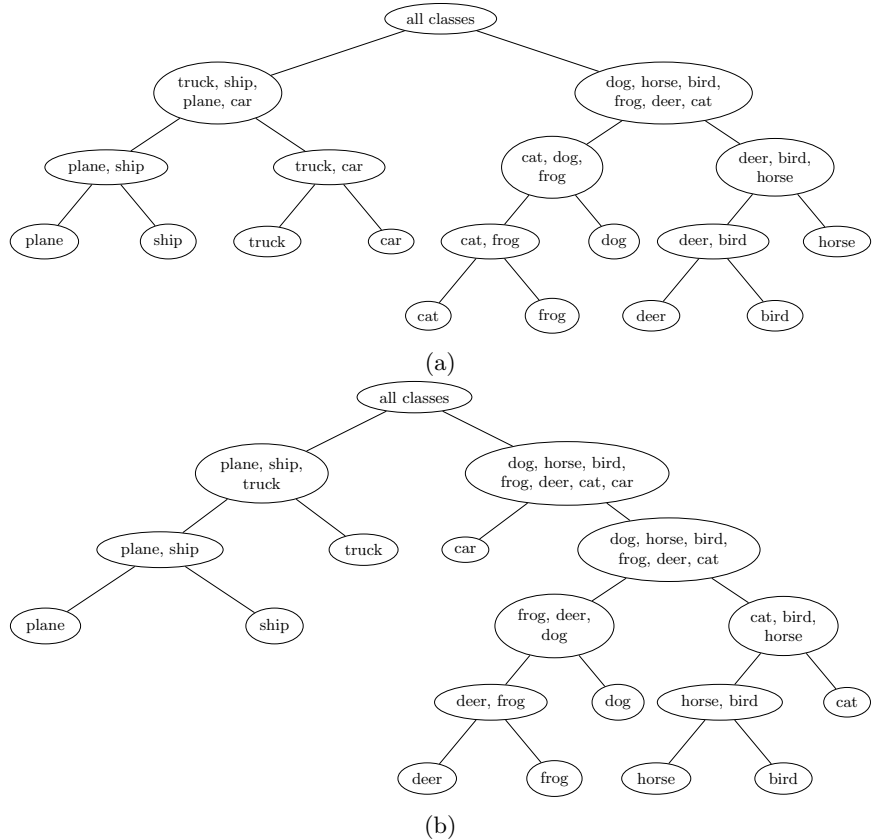

To test how well our method adapts to other base learners, we trained nested dichotomies with convolutional networks as the base learners to classify the CIFAR-10 dataset~\cite{krizhevsky2009learning}. Convolutional networks learn features from the data automatically, and perform well on high dimensional, highly correlated data such as images. We implemented the nested dichotomies and convolutional networks in Python using Lasagne~\cite{lasagne}, a wrapper for Theano~\cite{theano2,theano1}. The convolutional network that we used as the base learner is relatively simple; it has two convolutional layers with 32 $5\times5$ filters each, one $3\times3$ maxpool layer with $2\times2$ stride after each convolutional layer, and one fully-connected layer of $128$ units before a softmax layer.

As discussed in Section~\ref{sec:advantages}, the centroids for a dataset like CIFAR-10 appear to not be very descriptive, and as such, we expect NDBC with convolutional networks as the base learner to produce class splits that are not as well founded as those in RPND. We present a visualisation of the NDBC produced from the CIFAR-10 dataset, and an example of a nested dichotomy built with random-pair selection (Fig.~\ref{fig:cnn_trees}). We can see that both methods produce a reasonable dichotomy structure, but there are some cases in which the random-pair method results in more intuitive splits. For example, the root node of the RPND splits the full set of classes into the two natural subsets (vehicles and animals), whereas the NDBC omits the `car' class from the left-hand subset. Two pairs of similar classes in the animal subset -- `deer' and `horse', and `cat' and `dog' -- are kept together until near the leaves in the RPND, but are split up relatively early in the NDBC. Despite this, the accuracy and runtime of both methods were comparable. Of course, the quality of the nested dichotomy under random-pair selection is dependent on the initial pair of classes that is selected. If two classes that are similar to each other are selected to be the initial random pair, the tree can end up with splits that make less intuitive sense. 

\section{\label{sec:conclusions}Conclusion}
In this paper, we have proposed a semi-random method of class subset selection in ensembles of nested dichotomies, where the class selection is directly based on the ability of the base classifier to separate classes. Our method non-deterministically produces an easily separable class-split, which not only improves the accuracy over random methods for a single classifier, but also for ensembles of nested dichotomies. Our method also outperforms other non-random methods when nested dichotomies are used in a bagged ensemble and an ensemble boosted with MultiBoost, and otherwise gives comparable results. 

In the future, it would be interesting to explore selecting several random pairs of classes at each node, and choosing the best of the pairs to create the final class subsets. This will obviously increase the runtime, but may help to produce more accurate individual classifiers and small ensembles. We also wish to explore the use of convolutional networks in nested dichotomies further.

\subsubsection*{Acknowledgements.}
This research was supported by the Marsden Fund Council from Government funding, administered by the Royal Society of New Zealand. The authors also thank NVIDIA for donating a K40c GPU to support this research.

\bibliography{test}
\bibliographystyle{splncs03}

\end{document}